\documentclass[12pt]{article}

\usepackage[margin=0.75in]{geometry}
\usepackage{graphicx}

\usepackage{amsmath, amsthm, amssymb}

\usepackage[hyphens, spaces, obeyspaces]{url}
\usepackage[dvipsnames]{xcolor}
\definecolor{darkblue}{rgb}{0, 0, 0.5}
\usepackage[colorlinks, allcolors=darkblue, breaklinks=True]{hyperref}

\usepackage{tikz}
\usetikzlibrary{shapes, fit}

\usepackage[labelfont=bf,font=footnotesize,skip=1pt]{caption,subcaption}

\usepackage[T1]{fontenc}
\usepackage{newtxtext}

\usepackage{apacite}
\usepackage{doi}

\usepackage[nameinlink, noabbrev]{cleveref}

\usepackage{rotating}
\usepackage{bm}
\usepackage{lipsum}
\usepackage{booktabs}
\usepackage{siunitx}
\usepackage{threeparttable}
\usepackage{tablefootnote}
\usepackage{etoolbox}
\usepackage{titlesec}
\usepackage{multirow}
\usepackage{fancyhdr}
\usepackage{caption, subcaption}
\usepackage{titling}
\usepackage{blkarray}

\titlespacing{\section}{0pt}{14pt }{8pt}
\titlespacing{\subsection}{0pt}{8pt }{8pt}
\titlespacing{\subsubsection}{0pt}{8pt }{8pt}

\makeatletter
\patchcmd{\ttlh@hang}{\parindent\z@}{\parindent\z@\leavevmode}{}{}
\patchcmd{\ttlh@hang}{\noindent}{}{}{}
\makeatother

\titleformat{\section}
  {\normalfont \bfseries \fontsize{12.5pt}{12pt}\selectfont}
  { \thesection.}{6pt}{}

\titleformat{\subsection}
  {\normalfont \itshape \fontsize{12.5pt}{12pt}\selectfont}
  {\thesubsection}{6pt}{}

\titleformat{\subsubsection}
  {\normalfont \itshape \fontsize{12.5pt}{12pt}\selectfont}
  {\thesubsubsection}{6pt}{}

\makeatletter % access to internal commands
\renewcommand{\@seccntformat}[1]{\csname the#1\endcsname.\;}

\makeatother

\newcommand{\figtext}[1]{
	\captionsetup{justification=justified,
				singlelinecheck=false,
				font=footnotesize}
	\caption*{\hangindent=3.1em #1}
	}
\newcommand{\fignote}[1]{\figtext{\emph{Note:~}~#1}}

\newtheoremstyle{mystyle}
  {\topsep}
  {\topsep}
  {\itshape}
  {}
  {\scshape}
  {.}
  {.5em}
  {}
\theoremstyle{mystyle}

\renewenvironment{abstract}
               {\list{}{\setlength{\leftmargin}{.75in} \setlength{\rightmargin}{\leftmargin}}%
                \item[]{}\relax}
               {\endlist}

\renewcommand{\P}{\mathbb{P}}
\newcommand{\E}{\mathbb{E}}

\begin{document}

%%%%%%%%%%%%%%%%%%%
% Title page

% Title
\newcommand{\mytitle}{\uppercase{Machine learning classifiers do not improve the prediction of academic risk: Evidence from Australia}}
\pretitle{\begin{center}
	    	\fontsize{14pt}{16pt}\selectfont
		\bfseries}
\title{{\mytitle}}
\posttitle{\end{center}
		\vspace{-5mm}}
%%%%%%%%%%%%%%%%%%%

% Authors
\preauthor{\begin{center}
		    \fontsize{14pt}{16pt}\selectfont
		   }
\author{\small
{Sarah Cornell-Farrow\\
School of Economics, University of Adelaide, 10 Pulteney St, Adelaide, 5005, SA, Australia\\
\href{mailto:sarah.cornell-farrow@adelaide.edu.au}{sarah.cornell-farrow@adelaide.edu.au}
\\ \bigskip
Robert Garrard\thanks{}\\
CSIRO, Land \& Water, 41 Boggo Rd, Dutton Park, 4102, QLD, Australia\\
\href{mailto:robert.garrard@csiro.au}{robert.garrard@csiro.au}
}
}\postauthor{\end{center}}
% To put \thanks{} use \footnotemark[1]
%%%%%%%%%%%%%%%%%%%%%

% Date
\date{}

%%%%%%%%%%%%%%%%%%%%%
% Abstract

\twocolumn[
  \begin{@twocolumnfalse}
\maketitle

\vspace{-15mm}

    \begin{abstract}
      {
\fontsize{11pt}{12pt}\selectfont
\hspace{1em}
Machine learning methods tend to outperform traditional statistical models at prediction. In the prediction of academic achievement, ML models have not shown substantial improvement over logistic regression. So far, these results have almost entirely focused on college achievement, due to the availability of administrative datasets, and have contained relatively small sample sizes by ML standards. In this article we apply popular machine learning models to a large dataset ($n=1.2$ million) containing primary and middle school performance on a standardized test given annually to Australian students. We show that machine learning models do not outperform logistic regression for detecting students who will perform in the `below standard' band of achievement upon sitting their next test, even in a large-$n$ setting. 


\vspace{-1mm}
\\

	}
    \end{abstract}
  \end{@twocolumnfalse}
]

{
  \renewcommand{\thefootnote}
    {\fnsymbol{footnote}}
    
\footnotetext[1]{{This research is supported by an Australian Government Research Training Program (RTP) Scholarship. Thanks must also be given to the Australian Curriculum, Assessment and Reporting Authority for the provision of the data utilised by this study. The authors would like to thank Foivos Diakogiannis and Airong Zhang for their helpful comments and suggestions.}}

\footnotetext[2]{Jupyter notebooks containing R codes and outputs for this paper as well as instructions on how to obtain the data set may be found at \href{https://github.com/RobGarrard/A-Machine-Learning-Approach-for-Detecting-Students-at-Risk-of-Low-Academic-Achievement}{Github.com/RobGarrard}}

}

%%%%%%%%%%%%%%%%%%%%%%%%
% Misc
% Switch back to numbered footnotes
\renewcommand{\thefootnote}{\arabic{footnote}}
% Page header
\pagestyle{fancy}
%\fancyhead[HCE]{\textsc{\mytitle}}
\fancyhead[HCO]{\textsc{\uppercase{S. Cornell-Farrow and R. Garrard}}}
\lhead{}
\rhead{}
\renewcommand{\headrulewidth}{0pt}

%%%%%%%%%%%%%%%%%%%%%%%%%%%%%%%%%%%%%%%%%%%%%%%%%
\interfootnotelinepenalty=10000

\section{Introduction}
\label{sec:Introduction}

\noindent
Educational achievement is closely linked to an individual's economic well-being as well as a nation's standard of living \shortcite{Barro1991}. In Australia, the \textit{Report of the Review to Achieve Educational Excellence in Australian Schools} \shortcite{2gonski} highlighted the decline in student outcomes across the past twenty years when compared to other OECD countries. Not only are a large number of Australian students achieving poor outcomes, but there is a significant disparity in achievement within the same age groups and classrooms. 'One-size-fits-all' policy approaches have not been successful in bringing under-achieving students up to minimum standards. In order to support these students in reaching their full potential, it may be necessary to tailor policies and teaching strategies to individual students who are at risk of low achievement. Since schools, especially public schools, tend to face binding resource constraints, it is essential to detect `at risk' students early and with high precision.

In this setting, the goal is not to conduct causal inference on a coefficient of interest
%, as for usual econometric problems,
but to produce a purely predictive model that classifies students with high sensitivity and specificity. Machine learning (ML) methods tend to greatly outperform traditional statistical models \shortcite{Friedman2001}. For example, when attempting to classify images of handwritten digits correctly, standard logistic regression achieves an accuracy of around 70\%, whereas state-of-the-art neural networks obtain 99.79\% accuracy \shortcite{Wan2013}.

% ML estimators perorm well by exploiting the bias-variance tradeoff, in which the estimator is permitted to be biased in exchange for a large reduction in its variance, as well as incorporating non-linearities in a tractable way.

%%add something about that the collection of standardised test score data in Australia provides an interesting case study to test the performance of machine learning techniques with large-scale real world data. (or analysis of large-scale administrative data sets) (given that the journal is case studies)

The application of machine learning methods to education data has been referred to as Educational Data Mining \shortcite{Romero2007}. Its use in the prediction of academic performance has been predominantly in the higher education context \shortcite{Vandamme2007, Kotsiantis2012, Yadav2012, Jishan2015} largely due to the availability of administrative data sets collected by universities.\footnote{See \shortciteA{Shingari2017} for a recent review.} Interestingly, when ML models are benchmarked against logistic regression, they show no or unsubstantial improvement \shortcite{Kotsiantis2004, Cortez2008, HUANG2013, Gray2013}. This could be because the conditional expectation function in the predictors used in these studies does not exhibit significant non-linearities, or could  be that the sample sizes used are too small for the non-linearities to be learned.

In this article we exploit a large data set containing scores on the Australian National Assessment Program - Literacy and Numeracy (NAPLAN); a standardized literacy and numeracy test sat by all students in Australian schools in grades 3, 5, 7, and 9. This data set contains raw scores for all students who sat the test in the years 2013 and 2014, as well as administrative data on students' individual- and family-level characteristics. In total, the data set contains observations on 2.2 million unique students. Students are labeled into two classes, `At Standard' and `Below Standard', according to whether or not their score meets minimum achievement standards as determined by the Australian Curriculum, Assessment, and Reporting Agency (ACARA). This is done for two learning areas: literacy and numeracy. We split students into those in grade 3, for whom this would be their first time sitting NAPLAN, and students in grades 5 and above, for whom their previous achievement on NAPLAN may be used as a predictor.

We train a set of popular machine learning classifiers with standard logistic regression serving as a benchmark. We measure model performance using area under the ROC curve (AUC) and find that none of the machine learning models outperform logistic regression. To the best of our knowledge, this is the first paper to apply machine learning methods to the prediction of primary and middle school student achievement in a `big data' (large $n$) setting.

%%%where did our stuff about how the trees might be valuable for the policymaker, even if they don't improve predictive performance, go? we discuss this in the results so I would probably bring it up here too to increase the value-add of what we've done. or is it because they were too simple?

While machine learning methods have great potential for improving the analysis of economic data, the results of this study show that large data sets and state-of-the-art algorithms do not automatically imply better predictions.
%The rest of this article proceeds as follows. \Cref{sec:data} describes the data set. \Cref{sec:methods} describes preprocessing of the %data and the methods used to build the set of classifiers. \Cref{sec:results} presents the results, and \cref{sec:discussion} closes with a %discussion. 

%%%%%%%%%%%%%%%%%%%%%%%%%%%%%%%%%%%%%%%%%%%%%%%%%%%%%%%%%%
 
\section{Data}
\label{sec:data}
\noindent
NAPLAN is a set of standardized literacy and numeracy tests sat by all students in Australia in grades 3, 5, 7, and 9, in both the government and non-government schooling sectors. It is described as providing ``the measure through which governments, education authorities and schools can determine wh-ether or not young Australians are meeting important educational outcomes.''\footnote{See \href{http://www.nap.edu.au}{nap.edu.au}} The tests cover five learning areas known as `test domains': Reading, Writing, Spelling, Grammar and Punctuation, and Numeracy. The tests are designed to assess student performance relative to the Australian federal curriculum.

We use individual student-level data from the NAPLAN reading and numeracy tests administered in 2013 and 2014. The NAPLAN reading test involves a reading comprehension-style test on a range of texts, including imaginative, persuasive, and informative. The questions are designed to test knowledge and interpretation of English language in context. The NAPLAN numeracy tests assess students on their performance in mathematics, namely number and algebra, measurement and geometry, and statistics and probability.\footnote{Samples of these tests may be found at \url{https://www.nap.edu.au/naplan/the-tests}.} 

The data set contains 2,235,804 unique student IDs, who are attending 9,250 different schools in both the public and private sectors in all states across Australia. All individuals in the data set are unique, as a student sitting NAPLAN testing in 2013 would not sit the test in 2014, and vice versa. The individual scores for each student in each domain are collected by the Australian Curriculum Assessment and Reporting Authority (ACARA), alongside student background information which is collected by schools from students' parents or carers via enrolment forms. \Cref{tab:variablelist} provides a detailed summary of the variables contained in the data set.

For each testing domain in each year, ACARA determines achievement bands to classify a student's level of achievement based on what particular skills they can perform, \textit{e.g.}, addition of simple numbers, understanding of probability, \textit{etc.}. For each year level, students in the lowest two bands of achievement are deemed to be achieving `below minimum standards'. Given the raw scores for each student available in the data, we have classified each student into their relevant achievement band following the cut-off scores published by ACARA.\footnote{\url{nap.edu.au/results-and-reports/how-to-interpret/score-equivalence-tables}} Using these bands, we have labeled each student as either `At Standard' or `Below Standard' for reading and numeracy respectively. Within this sample, approximately 16-17\% of students are achieveing below standard performance in each learning domain.

In addition to each student's score in the 2013/14 NAPLAN cycle, the data also contains the score each student received in their previous testing cycle, two years earlier. This data is only present for students in grades 5 and above, as grade 3 is the first year NAPLAN is administered. In order to exploit this presumably strong predictor, we split the data set into students in grade 3, and students in grades 5 and above. We use these raw scores to construct a dummy variable for whether the student was `Previously At Standard' or `Previously Below Standard'.

%%%%%%%%%%%%%%%%%%%%%%%%%%%%%%%%%%%%%%%%%%%%%%%%%%%%%%%%
\section{Methods}
\label{sec:methods}

\noindent
Our objective is to predict whether or not a student will perform in the `Below Standard' band upon sitting their next NAPLAN. 

We split the data set into two subsets: students in grade 3, for whom this is their first time sitting NAPLAN; and students in grades 5, 7, and 9, for whom their NAPLAN score in the previous cycle may be used as a predictor. Within each data set, we remove any rows that contain missing data. For each data set (grade 3, $n=345,817$; and grades 5+, $n=886,392$) we obtain a random two thirds/one third split for training and test sets respectively. We stratify the random samples such that the proportion of students not meeting minumum standards is preserved in the training and test sets respectively. We do this for each of the response variables (reading and numeracy), resulting in each classifier being estimated separately on 4 data sets (reading/numeracy, grade 3/grade 5+) and evaluated on corresponding test sets.

In what follows, let $\left(y_{i}, \mathbf{x}_{i}^{\prime}\right)_{i=1}^{n}$ be a set of $n$ iid observations of a dependent variable $y_{i} \in \{0, 1\}$ with a corresponding vector of $p$ predictors $\mathbf{x}_{i}^{\prime} \in \mathcal{X}$ in a feature space $\mathcal{X}$.

\subsection{Classifiers}

\noindent
The task of a binary classifier is to predict whether or not an observation belongs to class 0 or 1 as a function of the predictors, $\mathbf{x}_{i}^{\prime}$. This is often accomplished by producing a predicted `probability' that the dependent variable belongs to the positive class, $\hat{y}_{i} \in [0, 1]$, and classifying the observation to that class if $\hat{y}_{i} \geq 0.5$.

A classifier's parameters are chosen such that the classifier's predictions minimize a specified loss function, such as mean square error or cross-entropy. Since we have a binary classification problem where one of the classes (`At Standard') is more common than the other (`Below Standard'), we use the weighted binary cross-entropy loss function.
%\begin{equation}
\begin{multline}
\mathcal{L}_{CE} =  \\
 -\frac{1}{n} \sum_{i} w_{i}\{y_{i} \log(\hat{y}_{i}) + (1-y_{i}) \log(1-\hat{y}_{i})\} \label{eqn:loss}
\end{multline}
%\end{equation}
Where each $w_{i}$ denotes a weight allowing some observations to contribute more or less to the loss function than others. Since the `At Standard' class is around 6 times more common than the `Below Standard' class, the loss function is weighted so that misclassifications of a `Below Standard' student incur around 6 times more cost. That is, $w_{i} = 6$ if the student performs below standard ($y_{i} = 1$), and $w_{i} = 1$ if the student performs at standard ($y_{i} = 0$).

Each model we consider gives a different functional form for generating $\hat{y}_{i}$ from the predictors. Unless otherwise stated, we choose the model parameters to minimize \cref{eqn:loss}. We estimate the following types of classifier.\footnote{In addition to these classifiers, other common choices include k-nearest neighbors and support vector machines. However, computation of these classifiers does not scale well with number of predictors and sample size respectively, and so were omitted from this study.}

%%%%%%%%%%%%%%%%%
\subsubsection{Logistic Regression}

\noindent
We use plain logistic regression as a benchmark for the ML classifiers. Logistic regression models the probability of belonging to the positive class as
\begin{equation}
\label{eqn:logistic}
\log \left( \frac{\P(y_{i} = 1)}{\P(y_{i} = 0)} \right) = \beta_{0} + \beta_{1} x_{1, i} + \dots + \beta_{p} x_{p, i}
\end{equation}
from which predicted values, $\hat{y}_{i} = \hat{\P}(y_{i} = 1)$, can be extracted by substituting in the predictor values and rearranging.
\smallskip

%%%%%%%%%%%%%%%
\subsubsection{Elastic Net}

\noindent
The elastic net \shortcite{Zou2005} is a hybrid of the lasso \shortcite{Tibshirani1996} and ridge regression \shortcite{Hoerl1970} making it a shrinkage estimator. Shrinkage estimators attempt to exploit the bias-variance tradeoff by introducing a small amount of bias through shrinking estimated coefficients towards zero. Shrinkage estimators are especially popular in high-dimensional regression, where $p >> n$, as they can perform automatic model selection by setting many coefficients exactly to zero. The elastic net has been used in genetic microarray studies for identifying genes associated with cancer risk \shortcite{Zou2005, Li2010, Li2013}.

The elastic net uses an identical functional form for generating predicted values as logistic regression but uses an alternative loss function for estimating the coefficients. Its loss function takes the form
\begin{equation}
\mathcal{L}_{EN} = \mathcal{L}_{CE} + \lambda\left( \alpha||\bm{\beta}||_{1} +(1-\alpha)||\bm{\beta}||_{2}^{2}\right)
\end{equation}
\noindent
where $\lambda$ is a tuning parameter determining the strength of the combination $\ell_{1}$ (lasso) and $\ell_{2}$ (ridge) penalty. Setting $\lambda = 0$ retrieves the unpenalized logistic regression, while setting $\lambda$ sufficiently large forces all coefficients to be zero. The parameter $\alpha$ controls the relative strength of the lasso and ridge penalties. The lasso penalty forces some of the coefficients to zero, performing model selection; while the ridge penalty forces correlated predictors to be assigned similar coefficients. 

We impose that $\alpha = 0.5$, such that the lasso and ridge penalties have equal weight and select $\lambda$ through 10-fold cross-validation.\footnote{The mixing parameter, $\alpha$, may also be tuned. However, this option is not included in the glmnet package. We re-estimated the classifiers with $\alpha= 0.1$ and $\alpha  = 0.9$ with no change to the results.} 
\smallskip

%%%%%%%%%%%%%%
\subsubsection{Decision Tree}

%%%%%%%%%
\begin{figure*}[t!]
	\begin{subfigure}[b]{0.47\textwidth}
		\includegraphics[width=\textwidth]{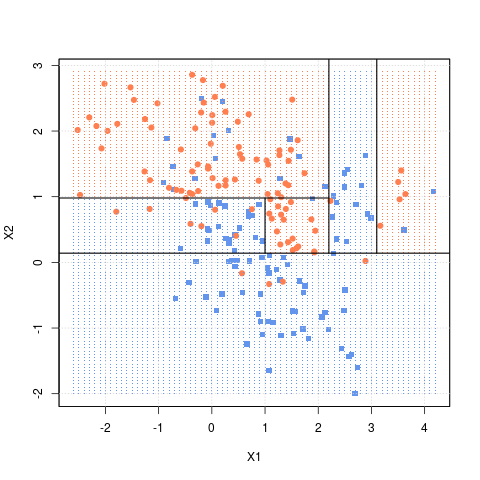}
		\caption{Feature space partitioned into regions. Squares denote $y_{i} = 0$, circles denote $y_{i} = 1$.}
		\label{subfig:rectangles}
	\end{subfigure}
	\hfill
	\begin{subfigure}[b]{0.47\textwidth}
		\includegraphics[width=\textwidth]{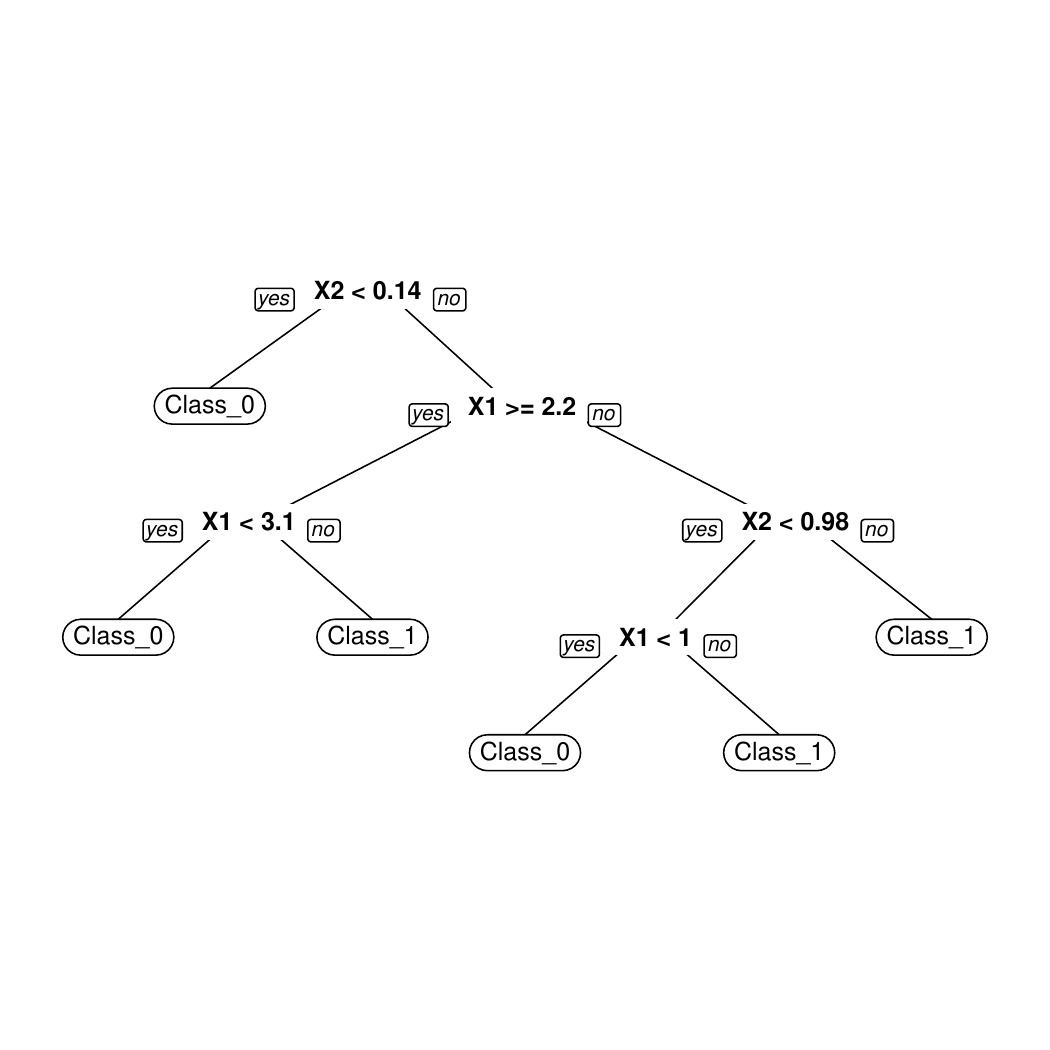}
		\caption{Decision tree predicting an observation to be in class 1 if $\hat{y}_{i} \geq 0.5$ and class 0 otherwise.}
		\label{subfig:tree_example}
	\end{subfigure}

\caption{Example of a binary classification tree. Feature space contains two continuous predictors, $X1$ and $X2$. Decision tree sequentially splits feature space into rectangles. Observations are classified according to class frequencies within each rectangle.}
\label{fig:tree_example}
\end{figure*}
%%%%%%%%%

\noindent
Classification and Regression Trees (CART) are a class of non-linear models that are tractable, flexible, and highly interpretable \shortcite{Breiman2017}. Decision trees can be visualized in a tree diagram representing a set of `if-then' conditions on the predictors, making them easily understood and applied by humans. Further, trees can naturally handle mixed data types and require only small modification to switch between a classification tree and a regression tree.

Let $R_{1}, \dots, R_{M} \subset \mathcal{X}$ be a set of regions that partition the feature space; i.e., $R_{i} \cap R_{j} = \emptyset$ for $i\not = j$ and $\cup_{i} R_{i} = \mathcal{X}$. Let $N_{m} = \sum_{i} I(\mathbf{x}_{i} \in R_{m})$ be the number of observations that fall in $R_{m}$, $m = 1, \dots, M$. Given such a partition, for an observation $i$ such that $\mathbf{x}_{i}^{\prime}$ falls in region $R_{m}$, define
\begin{equation}
\label{eqn:tree_classification}
\hat{y}_{i} = \frac{1}{N_{m}} \sum_{\mathbf{x}_{i}^{\prime} \in R_{m}} I\left(y_{i} = 1\right)
\end{equation}
such that the predicted probability of $i$ being in class 1 is the proportion of observations in $R_{m}$ belonging to class 1. 

Given this rule, ideally we would choose a set of regions in order to minimize our loss function, however this is in general computationally infeasible. Instead, a greedy algorithm is used that in each step chooses a predictor and splits the feature space in two at some value of that predictor. At the beginning of the algorithm, suppose we choose predictor $j$ and some threshold value for that predictor, $s$. These values can be used to split the feature space into two regions:
\begin{align}
R_{1} = \{\mathbf{x}_{i}^{\prime} \in \mathcal{X} \, | \, x_{i, j} \leq s\}\\
R_{2} = \{\mathbf{x}_{i}^{\prime} \in \mathcal{X} \, | \, x_{i, j} > s\}
\end{align}
Within these regions, observations are classified according to \cref{eqn:tree_classification}. So we may choose predictor $j$ and splitting value $s$ to minimize the loss function. The algorithm can then be repeated at each step within each newly created region.

Since each region is determined by a series of sequential splits, we can represent the classifier by a tree diagram in which each node is a predictor and the two branches emanating from the node represent which side of the split the value of that predictor falls on. How large the tree grows is a tuning parameter that can be cross-validated.

\Cref{fig:tree_example} illustrates an example of a classification tree.

\smallskip
%%%%%%%%%%%%%%
\subsubsection{Random Forest}

\noindent
The predictive performance of a decision tree can be improved at the expense of interpretability by forming an ensemble of trees. Bootstrap aggregated (bagged) decision trees can be formed by training separate decision trees on bootstrap samples of the data \shortcite{Breiman1996}. On different realizations of the data from each bootstrap sample, different decision rules can be learned by each tree. Averaging predictions across these trees serves to decrease prediction bias.

However, if a few of the predictors are particularly strong, the decision trees will have a tendency to use these predictors in the early splits on most of the bootstrap samples, leading to small variation in the rules learned by the trees. To combat this, \shortciteA{Breiman2001} suggested only allowing a random subset of predictors to be used on each bootstrap sample to promote variation in the rules learned by the trees.

In addition to the size of each tree being a tuning parameter we now also need to tune the bootstrap sample size, the number of predictors to randomly choose on each bootstrap, and the number of trees in the ensemble. Standard convention is to use bootstrap samples of same size as the data, $n$. Default values for the number of predictors to randomly sample for each tree are $\sqrt{p}$ for classification and $p/3$ for regression \shortcite[pp. 592]{Friedman2001}. Since random forests do not overfit as the number of trees increases, the number of trees can be as large as the modeler desires.

We train a random forest with out-of-the-box parameter choices for the bootstrap sample size and number of predictors selected in each sample. We train an ensemble with 500 trees, which keeps the compute time manageable given the size of our data set.
\smallskip

%%%%%%%%%%%%%%

\subsubsection{Neural Network}

\noindent

Neural networks are an especially popular machine learning tool because they can learn sophisticated non-linear functions when both sample sizes and number of predictors are large. Contrast this with decision trees, which learn relatively unsophisticated non-linear functions due to partitioning the feature space into rectangles, rather than more complicated shapes; and support vector machines or k-nearest neighbors, which become computationally infeasible for large $n$ and $p$. Neural networks have been used to forecast time series, translate text, and learn to play games. They are particularly useful in `deep learning' for computer vision, where they are used classify and label objects in images, automatically generate captions for images, and generate photorealistic images from random noise.

\begin{figure}[t!]
\centering
    \begin{tikzpicture}

% Inputs
    \node (A1) at (0,1) {$x_{1}$};
    \node (A2) at (0,0) {$\vdots$};
    \node (A3) at (0,-1) {$x_{k}$};
    
    % Middle node
    \node[circle, draw] (B) at (2,0) {};
    \node (b) at (2, -2) {$y = \sigma\left(\beta_{0} + \sum_{i=1}^{k} \beta_{i}x_{i}\right)$};
    
    \node (C) at (4,0) {$y$};

    % Connections
	\draw[->] (A1) -- (B);
	\draw[->] (A2) -- (B);
	\draw[->] (A3) -- (B);
	
	\draw[->] (B) -- (C);
	\end{tikzpicture}
	
\caption{A single neuron in a neural network. A vector of $k$ inputs enters the neuron, either from an input layer or from outputs of $k$ neurons in the previous layer. A linear combination of these inputs is taken using a set of weights, $\beta_{i}$, and a bias (intercept) term, $\beta_{0}$. This linear combination is mapped through a non-linear activation function. The output of this function is transmitted to neurons in the next layer.}
\label{fig:neuron}
\end{figure}
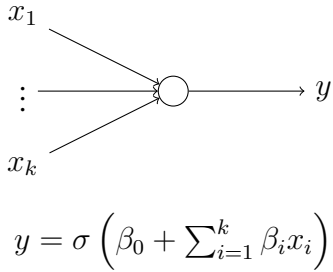

%%%%%%%%%%%
% Neural Net

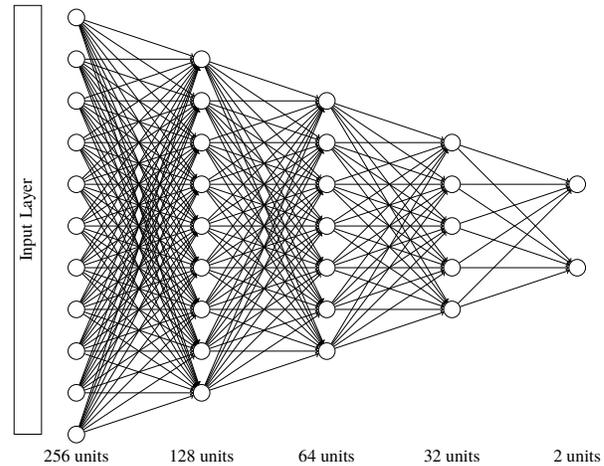
\begin{figure}
\centering

\resizebox{0.45\textwidth}{!}{
	\begin{tikzpicture}
%	\draw[step=1cm,gray,very thin] (0,0) grid (14,12);
	
	% Input layer
\node (a) [shape=rectangle,
		 draw,
         		 text width=10cm,
 	           rotate=90,
         		  minimum width=1cm,
	           anchor=north west,
     		 align=center] at (-1.5, 1) {Input Layer};

	% Hidden layer 1
	\foreach \i in {1, ..., 11} {
		\node[circle, draw] (A) at (0,\i) {};
		
		\foreach \j in {2, ..., 10}{
			\node[circle] (B) at (3, \j) {};
			\draw[->] (A) -- (B);
		}
	}

\node[align=center] at (0, 0.5) {256 units};

	% Hidden layer 2
	\foreach \i in {2, ..., 10} {
		\node[circle, draw] (A) at (3,\i) {};
		
		\foreach \j in {3, ..., 9}{
			\node[circle] (B) at (6, \j) {};
			\draw[->] (A) -- (B);
		}
	}

\node[align=center] at (3, 0.5) {128 units};

	% Hidden layer 3
	\foreach \i in {3, ..., 9} {
		\node[circle, draw] (A) at (6,\i) {};
		
		\foreach \j in {4, ...,8}{
			\node[circle] (B) at (9, \j) {};
			\draw[->] (A) -- (B);
		}
	}

\node[align=center] at (6, 0.5) {64 units};

	% Hidden layer 4
	\foreach \i in {4, ..., 8} {
		\node[circle, draw] (A) at (9,\i) {};
		
		\foreach \j in {5, 7}{
			\node[circle] (B) at (12, \j) {};
			\draw[->] (A) -- (B);
		}
	}
\node[align=center] at (9, 0.5) {32 units};

	% Output layer
	\foreach \i in {5, 7} {
		\node[circle, draw] () at (12,\i) {};
	}

\node[align=center] at (12, 0.5) {2 units};

	\end{tikzpicture}
}

\caption{Neural net architecture. The net contains 4 hidden layers containing 256, 128, 64, and 32 hidden units respectively. Each hidden unit uses a ReLU activation function with dropout regularization of 20\%. The output layer contains 2 units, corresponding to the probabilities of `At Standard' and `Below Standard', and uses softmax activation.}
\label{fig:neural_net}
\end{figure}

%%%%%%%%%%%%%%%%

% Talk about an individual neuron

Neural networks attempt to imitate the biological function of neurons and axons in a brain. The basic building block of a neural network is a neuron (or hidden unit). Neurons accept a vector of input values, perform a linear transformation on those values, apply an activation function to the output of that transformation, and output the result of the activation function (\cref{fig:neuron}). The coefficients (or weights) and intercept term (or bias) in this linear transformation are parameters of the neural network that must be estimated.

The application of an activation function is what grants the neural network non-linearity. The activation function attempts to emulate an action potential that decides whether or not a neuron in the brain `fires'; i.e., passes a large output value on to other neurons. Popular choices for the activation function are the sigmoid function, and the rectilinear function. 
\begin{align}
    &\sigma_{sigmoid}(x) = \frac{1}{1 + e^{-x}}\\
    &\sigma_{ReLU}(x) = x I(x > 0)
\end{align}
Neurons with a rectilinear activation function are called rectilinear units (ReLU). The neural network is built by connecting a set of neurons together such that the outputs of some neurons are the inputs of others.

In this article we use a popular neural network architecture called a multilayer perceptron (MLP). An MLP structures the neural network as an input layer, a set of hidden layers, and an output layer. The input layer does not contain any neurons and simply feeds the vector of predictors, $\mathbf{x}_{i}^{\prime}$, into each neuron in the first hidden layer. Hidden layers contain a set of neurons that are connected to the previous and subsequent layers, but share no connections within the same layer. There may be many hidden layers, or only one (as in the case of the Single Layer Perceptron), and each hidden layer may have a different number of neurons. The output layer contains a set of neurons whose output corresponds to the output of the function the neural network is estimating and tends to have a different activation function than hidden layers. For regression problems, the output layer is typically a single neuron with a linear activation function; and for a $k$-class classification problem, the output layer contains $k$ neurons, corresponding to the predicted probabilities than an observation is in class $k$, with a softmax activation function to normalize those probabilites such that they sum to one. 

\Cref{fig:neural_net} illustrates the MLP structure used here. We use 4 hidden layers with 256, 128, 64, and 32 neurons respectively. Since we are trying to classify a binary variable, the output layer of the neural net contains two units and a softmax activation function. The predicted probability that an observation belongs to the positive class, $\hat{y}_{i}$, corresponds to the output of one of these neurons.

The neural network's weights are estimated by minimizing the loss function using the backpropogation algorithm \shortcite{Rumelhart1985}. Formerly, common practice to avoid overfitting was to stop the training of the neural network early according to error on a validation set. However, with the invention of the \textit{dropout layer} \shortcite{Srivastava2014}, overfitting can be avoided while allowing the network to be trained to optimality. Dropout layers accomplish this by randomly preventing a proportion of neurons in each layer in each training batch from propogating their output forward into the next layer. This is somewhat analogous to random forests, where a random subset of predictors are used to build each decision tree. We regularize the network using dropout layers after each hidden layer, with dropout rate set to 20\%. The network is trained for 100 epochs.
\smallskip

\subsection{Model Evaluation}

\noindent
Binary classification models are often evaluated using accuracy (or balanced accuracy) on a test set. Accuracy is calculated as the proportion of correct predictions, where an observation is predicted to be in class 1 if $\hat{y}_{i} \geq 0.5$ and class 0 otherwise. However, in some settings we may wish to err on the side of a classifier being particularly sensitive or specific. For example, in scientific literature low sensitivity is tolerated in order to have high specificity, e.g., through the $p < 0.05$ convention for p-values, in an attempt to avoid false discoveries. Whereas in medical testing high sensitivity may be preferable to avoid failing to detect a serious medical condition. Similarly, in the early detection of academic risk, one might wish to favor sensitivity or specificity in a classifier. 

A range of sensitivity/specificity combinations can be achieved for a classifier by classifying an observation into class 1 if $\hat{y}_{i} \geq z$, for some $z \in [0, 1]$. The frontier of sensitivity/specificity combinations that a classifier can achieve by varying $z$ is called the Receiver Operating Characteristic (ROC) curve. In order to remain agnostic with respect to a desirable sensitivity or specificity, we measure model performance using area under the ROC curve (AUC) evaluated on a test set. 

Let $C_{1} = \{ i \, | \, y_{i} = 1\}$ be the set of observations not achieving minimum standards and $C_{2} = \{j \, | \, y_{i} = 0\}$ be the set meeting minimum standards, with $|C_{1}| = N_{X}$ and $|C_{2}| = N_{Y} = n - N_{X}$. For a given classifier, let $\{X_{i}\}_{i=1}^{N_{X}}$ be the set of predicted probabilities $X_{i} = \hat{\P}(y_{i} = 1)$ for each $i \in C_{1}$ and let $\{Y_{j}\}_{j=1}^{N_{Y}}$ be defined analogously for $j \in C_{2}$. 

Given some threshold $z \in [0, 1]$, the classifier predicts that an observation $i \in C_{1}$ belongs to the positive class if $X_{i} \geq z$, and analogously for $j \in C_{2}$. The classifier makes a correct prediction when $X_{i} \geq z$ or $Y_{i} < z$.

\begin{figure}[t!]
%	\centering
	\includegraphics[width=0.45\textwidth]{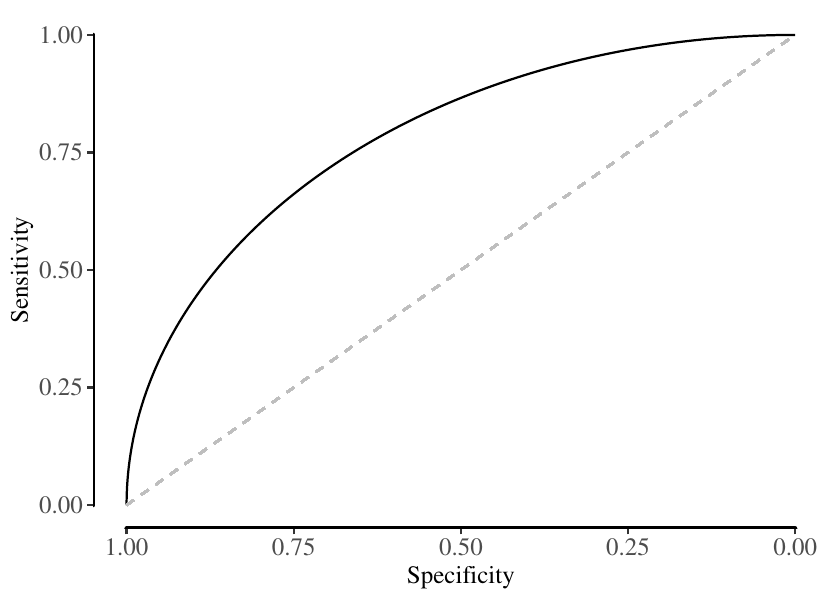}
\caption{Example of a ROC curve. Note that when the threshold $z = 0$, all observations are classified as the positive class giving sensitivity of 1. Similarly, when $z=1$, all observations are classified in the negative class.}
\label{fig:roc_example}
\end{figure}

The sensitivity and specificity of the classifier are defined as follows.
\begin{align}
\label{eqn:sensitivity}
&\text{sens}(z) = \frac{1}{N_{X}} \sum_{i \in C_{1}} I(X_{i} \geq z)\\
\label{eqn:specificity}
&\text{spec}(z) = \frac{1}{N_{Y}} \sum_{j \in C_{2}} I(Y_{i} < z)
\end{align}
The set of sensitivity/specificity pairs for $z\in[0, 1]$ defines the ROC curve. \Cref{fig:roc_example} illustrates. The area under the ROC curve (AUC) neatly summarizes the set of achieveable sensitivity/specificity combinations. An AUC of 1 is most desirable, corresponding to both a sensitivity and specificity of 1. One classifier with larger AUC than another is able to achieve a greater sensitivity for a given specificity, and vice versa.

\shortciteA{BAMBER1975} shows that when the area under the ROC curve is calculated with the trapezoidal rule, the estimator for AUC is equal to
\begin{equation}
\widehat{AUC}(X, Y) = \frac{1}{N_{X}N_{Y}} \sum_{j=1}^{N_{Y}} \sum_{i=1}^{N_{X}} \psi(X_{i}, Y_{j})
\end{equation}
where
\begin{equation}
\psi = 
	\begin{cases}
	1 \ \text{if} \ X < Y\\
	\frac{1}{2} \ \text{if} \ X = Y\\
	0 \ \text{if} \ X > Y
	\end{cases}
\end{equation}
In the case where $X$ and $Y$ are continuous, such that $\P(X = Y) = 0$, the AUC is equivalent to the Mann-Whitney U statistic, which is asymptotically normal with known variance. \shortciteA{BAMBER1975} uses a result from \shortciteA{noether1967} to estimate the variance of AUC when $X$ and $Y$ are not continuous. Using the formulation from \shortciteA{DeLong1988}, define
\begin{align}
\xi_{10} = \E[\psi(X_{i}, Y_{j})\psi(X_{i}, Y_{k})] - AUC^{2} \quad j \not = k \nonumber\\ 
\xi_{01} = \E[\psi(X_{i}, Y_{j})\psi(X_{k}, Y_{j})] - AUC^{2} \quad i \not = k \nonumber\\ 
\xi_{11} = \E[\psi(X_{i}, Y_{j})\psi(X_{i}, Y_{j})] - AUC^{2} \quad i \not = j
\label{eqn:covars}
\end{align}
The variance for the AUC estimate can then be expressed as
\begin{equation}
\label{eqn:var_estimate}
\text{Var}(\widehat{AUC}) = \frac{(N_{X}-1)\xi_{10} + (N_{Y} - 1)\xi_{01} + \xi_{11}}{N_{X}N_{Y}}
\end{equation}
This variance can be estimated by replacing population quantities in \cref{eqn:covars} with their sample analogues and confidence intervals for the AUC estimate can then be constructed according to asymptotic normality.

%%%%%%%%%%%%%%%%%%%%%%%%%%%%%%%%%%%%%%%%%%%%%%%%%%%%%%%%%%
%%%%%%%%%%%%%%%%%

\newcolumntype{L}[1]{>{\raggedright\let\newline\\\arraybackslash\hspace{0pt}}m{#1}}
\newcolumntype{C}[1]{>{\centering\let\newline\\\arraybackslash\hspace{0pt}}m{#1}}

\begin{table*}[!ht]
\centering
\caption{Performance metrics. Area under the curve is between 0 and 1, with 1 being most preferable.}
\label{tab:performance_metrics}

\begin{tabular}{L{4cm}  ccc C{1cm} ccc}
\toprule
\multicolumn{8}{c}{Grade 3}\\
\midrule
Classifier & \multicolumn{3}{c}{Literacy} && \multicolumn{3}{c}{Numeracy}\\
\midrule
& AUC && 95\% CI && AUC && 95\% CI\\
\cline{2-4} \cline{6-8}\\
Logistic& 0.722 && (0.717, 0.726) && 0.707 && (0.702, 0.712)\\
Elastic Net& 0.721 && (0.717, 0.726) &&  0.707 && (0.702, 0.712)\\
Decision Tree& 0.655 && (0.650, 0.660) && 0.668 && (0.663, 0.673)\\
Random Forest& 0.697 && (0.692, 0.702) && 0.681 && (0.676, 0.686)\\
Neural Net & 0.718  && (0.713, 0.722) && 0.700 && (0.695, 0.705)\\
\midrule
\multicolumn{8}{c}{Grade 5+}\\
\midrule
Classifier & \multicolumn{3}{c}{Literacy} && \multicolumn{3}{c}{Numeracy}\\
\midrule
& AUC && 95\% CI && AUC && 95\% CI\\
\cline{2-4} \cline{6-8}\\
Logistic& 0.839 && (0.837, 0.841)  &&  0.833 && (0.830, 0.835)\\
Elastic Net& 0.839 && (0.837, 0.841) && 0.832 && (0.830, 0.835)\\
Decision Tree& 0.767 && (0.765, 0.770) && 0.761 && (0.758, 0.763)\\
Random Forest& 0.828 && (0.826, 0.831) && 0.823 && (0.820, 0.825)\\
Neural Net& 0.833 && (0.831, 0.835) && 0.827 && (0.825, 0.830)\\
\bottomrule
\end{tabular}
\end{table*}

%%%%%%%%%%%%%%%%%%%

\section{Results}
\label{sec:results}

\begin{figure}[!p]
\centering

\begin{subfigure}[b]{.45\textwidth}
\includegraphics[width=\textwidth]{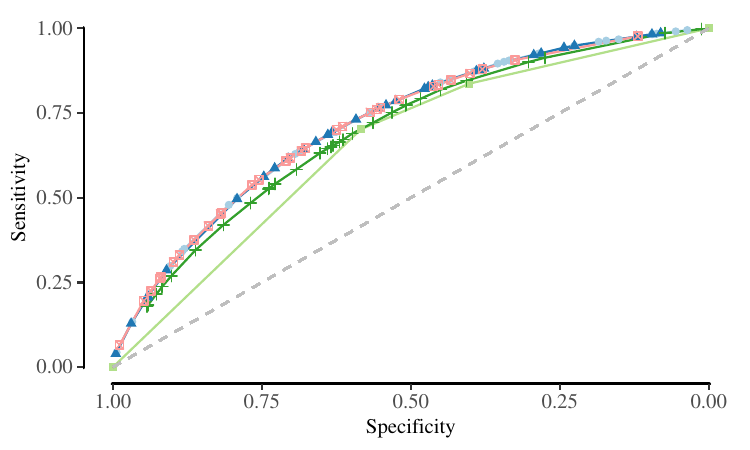}
\caption{Grade 3. Literacy.}
\end{subfigure}\hfill
\begin{subfigure}[b]{.45\textwidth}
\includegraphics[width=\textwidth]{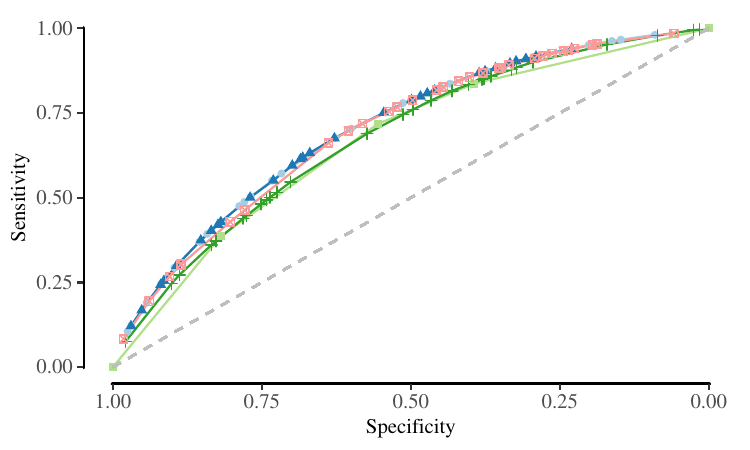}
\caption{Grade 3. Numeracy}
\end{subfigure}
\newline

\begin{subfigure}[b]{.45\textwidth}
\includegraphics[width=\textwidth]{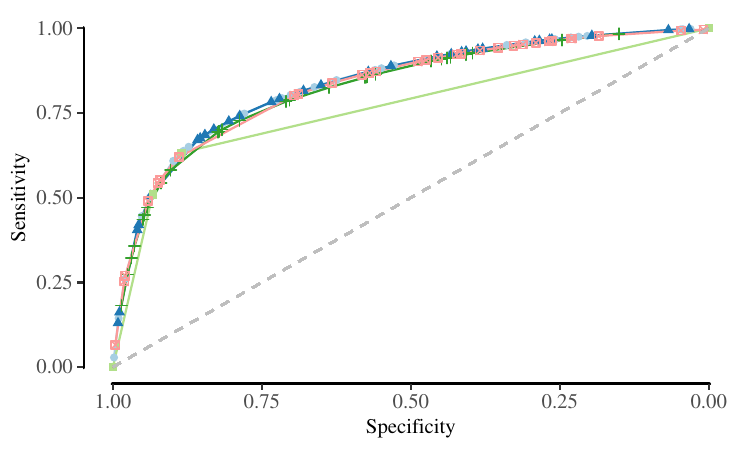}
\caption{Grade 5+. Literacy.}
\end{subfigure}\hfill
\begin{subfigure}[b]{.45\textwidth}
\includegraphics[width=\textwidth]{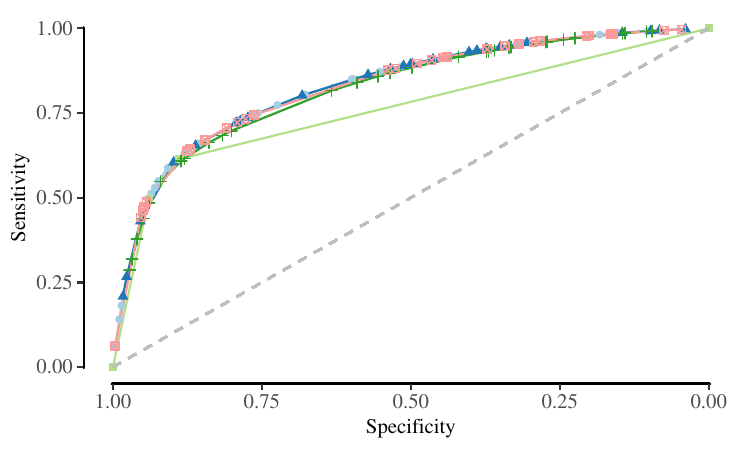}
\caption{Grade 5+. Numeracy.}
\end{subfigure}
\begin{subfigure}[b]{\textwidth}
\includegraphics[width=\textwidth]{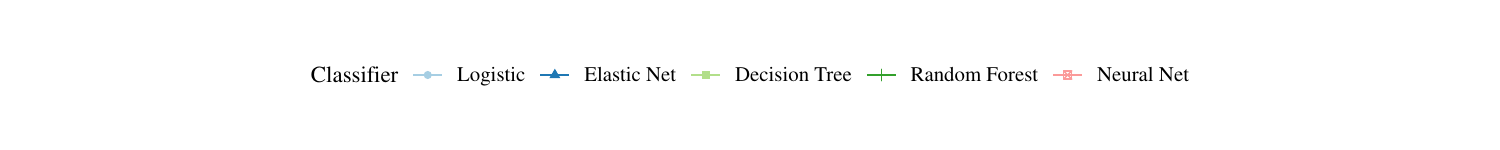}
\end{subfigure}

\caption{ROC curves. Each curve shows the frontier of sensitivity/specificity combinations each classifier achieves on a test set. An ideal classifier achieves both sensitivity and specificity of 1.}
\label{fig:ROC_curves}
\end{figure}

\noindent
ROC curves for each classifier are displayed in \cref{fig:ROC_curves}. \Cref{tab:performance_metrics} contains performance metrics for each model. We see that no ML model is able to outperform logistic regression in terms of AUC. The logistic classifier has the highest or equal highest AUC in all cases and in many cases the decision tree, random forest, and neural network perform statistically significantly worse. The elastic net matched the performance of the logistic, indicating that the penalty used to exploit the bias-variance trade-off did not improve predictive ability. Further, the decision tree and random forests significantly underperformed the logistic regression and elastic net.

The reason for ML not delivering substantial improvement in the academic performance setting possibly resides in the nature of common predictors used being categorical. Part of ML's power comes from its ability to tractably model non-linearities in the data. A lack of continuous predictors with substantial non-linearities being present in education data may explain part of this result. However, ML can still give substantial improvement even when most predictors are categorical. Decision trees (and hence random forests) and neural nets can naturally model complex interaction effects between categorical predictors. These results suggest that such interaction effects may not be present in the education setting either.

Since the grade 5 and above data contains past performance predictors, these models perform significantly better than their grade 3 counterparts. The logistic AUCs for the Grade 5+ predictions are 0.839 and 0.833 for literacy and numeracy respectively, compared to only 0.722 and 0.707 for the Grade 3 predictions. While this is to be expected, since the grade 5+ data set has an additional strong predictor, it is desirable to detect underachievement as early as possible due to its persistence.  

\begin{sidewaysfigure}[p]
\centering
	
	\begin{subfigure}[b]{0.45\textwidth}
		\centering
		\includegraphics[width=\textwidth, trim={0in, 1.5in, 0in, 1.5in}, clip]{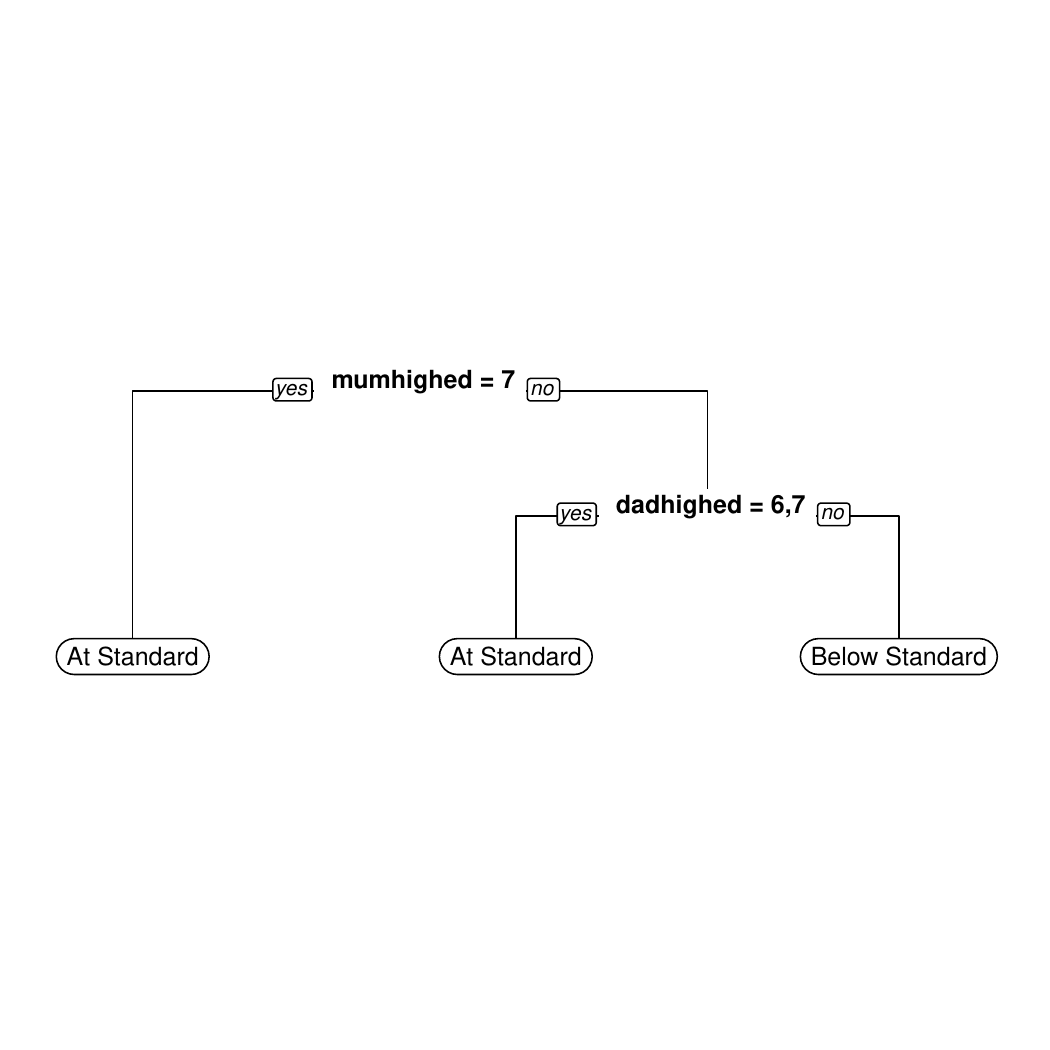}
		\caption{Literacy. Grade 3.}
	\end{subfigure}
	\begin{subfigure}[b]{0.45\textwidth}
		\centering
		\includegraphics[width=\textwidth, trim={0, 1.5in, 0, 1.5in}, clip]{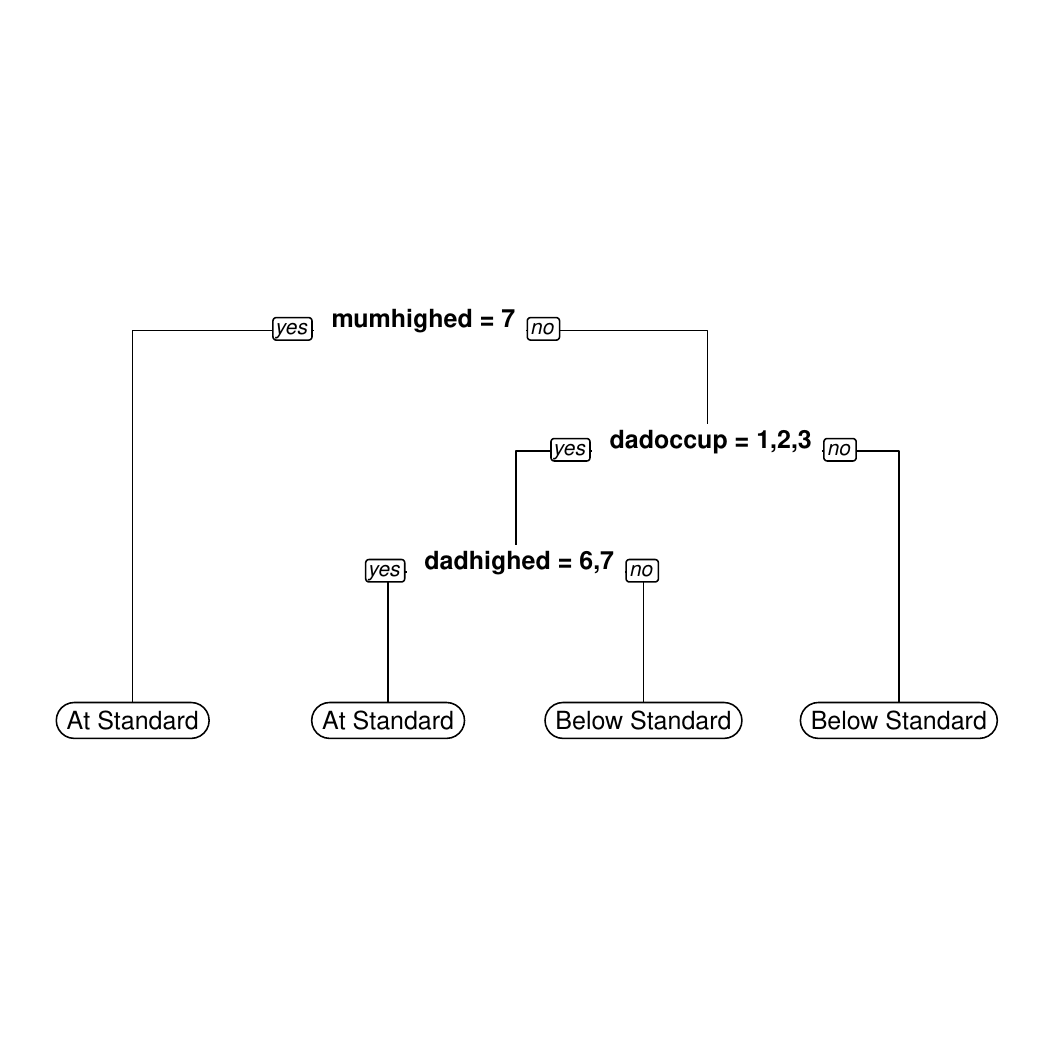}
		\caption{Numeracy. Grade 3.}
	\end{subfigure}

	\begin{subfigure}[b]{0.45\textwidth}
		\centering
		\includegraphics[width=\textwidth, trim={0in, 1.5in, 0in, 1.5in}, clip]{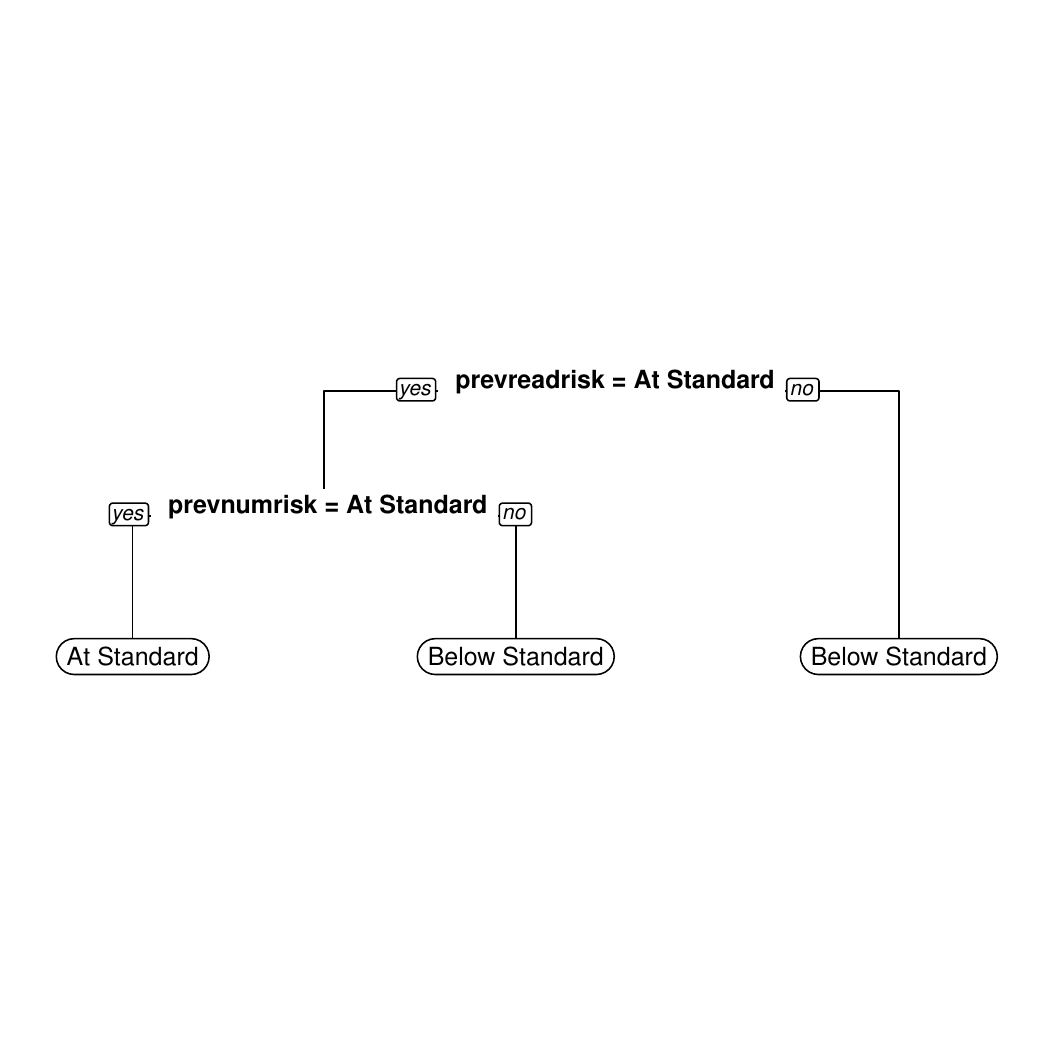}
		\caption{Literacy. Grade 5+.}
	\end{subfigure}
	\begin{subfigure}[b]{0.45\textwidth}
		\centering
		\includegraphics[width=\textwidth, trim={0, 1.5in, 0, 1.5in}, clip]{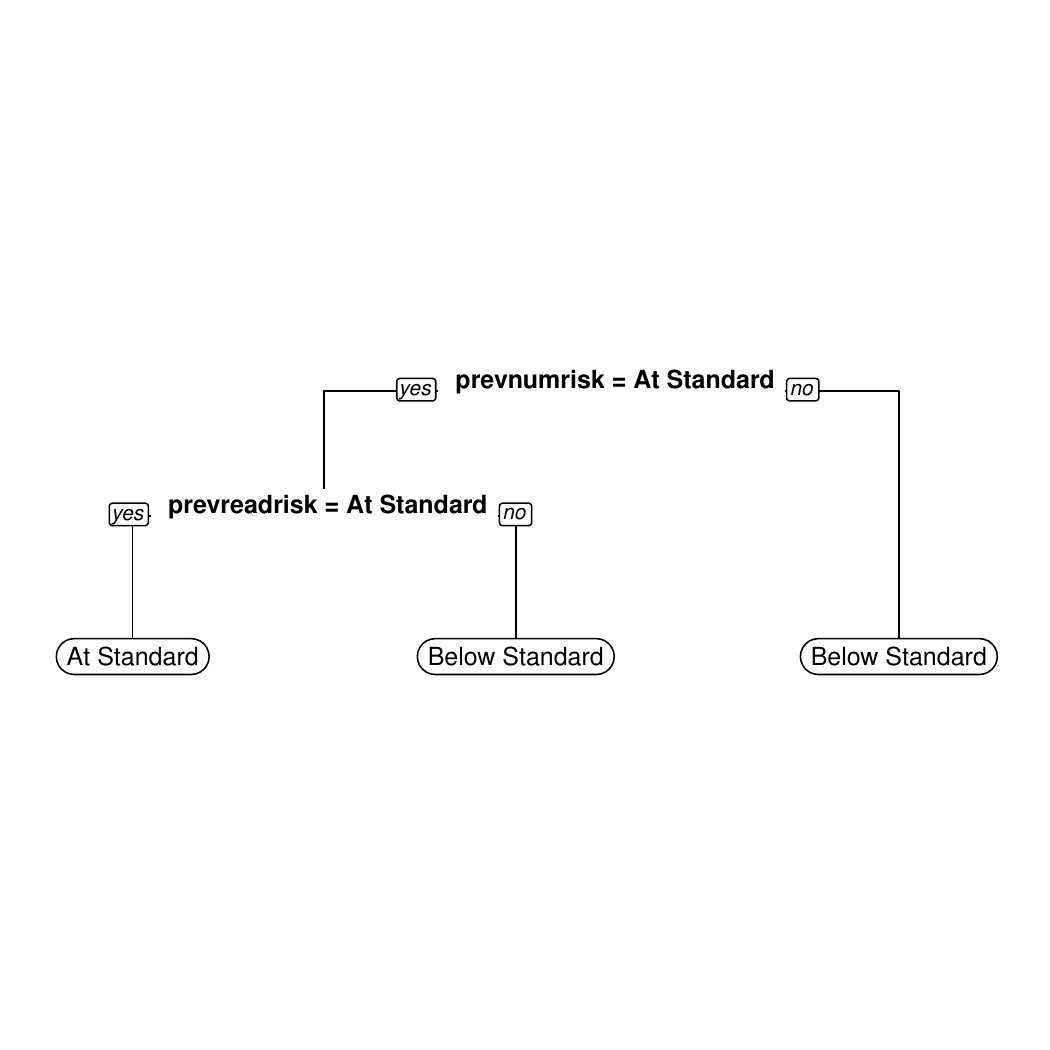}
		\caption{Numeracy. Grade 5+.}
	\end{subfigure}

\caption{Decison trees. See \cref{tab:variablelist} for variable descriptions.}
\label{fig:decision_trees}
\end{sidewaysfigure}

While the tree-based methods perform significantly worse than logistic regression, they give highly interpretable decision rules. These rules may be useful in terms of interpretation, as they provide simple rules or heuristics for detecting students at risk of poor performance. The potential advantage of such simplicity is that they could easily be learned and applied with minimal effort by school staff. \Cref{fig:decision_trees} displays.

When previous achievement on NAPLAN is available (i.e. for Grade 5+), the decision trees learn a very simple and intuitive decision rule. If a student was below standard on \textit{either} literacy or numeracy in the previous testing period, they are classified as at risk in \textit{both} subjects in the next period. This is of value, as it indicates that a school faced with a student that has performed below standard in numeracy in one period should not shift all of their resources towards improving only their math ability. In the United States, \shortciteA{Reback2008} highlighted that Texan schools reallocated resources towards students based on their performance on previous tests. We provide evidence that this may not be a useful approach, and instead a focus on overall improvement in all subject areas should be emphasised.

The decision trees for grade 3, for which achievement in the previous period is not available, provide some insight into the predictors of poor achievement in Australia. Categorising students as at risk or not depends completely on the education levels and occupation of their parents. For both literacy and numeracy, the first split is made regarding mother's higher education. If a student's mother holds a Bachelor's degree, a student will be predicted to meet minimum standards. However, if the mother does not hold a degree, their performance depends on the education of their father. If their father also does not have higher education qualifications, the student will be predicted not to meet minimum standards. If their father has either a Bachelor's degree, a diploma or an advanced diploma, the student will be predicted to meet minimum standards. In general terms, if \textit{at least one} of the student's parents has a Bachelor's degree, or their father has a diploma or an advanced diploma, a student is not at risk of poor performance on NAPLAN, in reading or numeracy. This finding matches much of the literature, which finds that family background strongly determines educational outcomes \shortcite{Cobb2012,Ford2013}. In fact, \shortciteA{Nicoletti2013} found that family background explains 44-55\% of the variation in test scores for students in England.

The results are the same for literacy and numeracy, save the addition of a decision rule regarding father's occupation for numeracy. If the father is employed, and is not employed in a 'category 4' job, a student will be predicted not at risk of not meeting minimum standards in the numeracy test. Category 4 encompasses machine operators, hospitality staff, assistants, labourers and related workers.

While these decision rules are simple and intuitive, they may be limited in their practical application due to the broad nature of each category. However, these results show the feasibility and potential for these kinds of models for predicting test score outcomes in future should more detailed data become available.

%%%%%%%%%%%%%%%%%%%%%%%%%%%%%%%%%%%%%%%%%%%%%%%%%%%%%%%%%%%%%%
\section{Discussion}
\label{sec:discussion}

\noindent
The machine learning models we estimated in the paper did not outperform standard logistic regression. Logistic regression was equal best in terms of the point estimate for model performance, so no ML model even managed to achieve a statistically insignificant improvement. 

A partial explanation for this could be the relative lack of numerical predictors, for which our data set contained only age.\footnote{We performed the analysis with the two previous NAPLAN achievement variables as numerical scores rather than dummies for `at standard' or `below standard' achievement with no change to the results.} Part of ML's ability to improve predictive power comes from using non-linear estimators of the conditional expectation function, whereas OLS and logistic regression are limited to estimating linear functions (with the log-odds ratio being linear in the case of logistic regression). Having few numerical predictors, or numerical predictors for which the conditional expectation function is close to linear could prevent ML models from performing better.

However, in combining predictors non-linearly, ML models should also be able to learn interaction effects if they are present. If one suspected that interaction effects were relevant in a prediction problem involving OLS or logistic regression, but did not know which interactions predicted well, one could saturate the model with all combinations of two variables. For $p$ predictors, this would involve adding on the order of $p^{2}$ additional predictors. For a large number of predictors or a small sample size, this would become a high-dimensional problem and one would have to resort to machine learning methods anyway, such as regularization or subset selection. Alternatively, one could use a model that can learn interaction effects without having to be given interaction predictors, such as decision trees and neural networks. For our data set, strong interaction effects appear not to be present.

Our result aligns with studies in the educational data mining literature that have benchmarked ML models against logistic regression in the prediction of academic performance \shortcite{Kotsiantis2004, Cortez2008, HUANG2013, Gray2013}, although while our study finds no improvement at all from ML, some of these studies find a slight increase in predictive power. These studies have primarily used college level administrative data and have had comparatively small sample sizes. We have shown that this result also holds at the primary and middle school level with a data set of over a million observations. In order for ML to gain traction in this setting, we would likely need data sets richer in numerical predictors, such as scores on formative assessment from classes taken in previous years. However, in the primary and middle school setting, this may not be feasible if letter grades are used more commonly than numeric scores. Additionally, model performance could be improved by collecting predictors that are time varying. Most of the predictors in our model are time invariant, except age, grade, and previous NAPLAN performance.\footnote{Employment status of the parents is time varying, but is only collected at the time of enrolment.}

Finally, any exercise in producing an early warning model for academic risk needs to be mindful of possible data set contamination from unobserved interventions. Presumably the teachers of students in a data set are aware when one of their students has not obtained minimum achievement and may attempt to mobilize school resources to correct for a student that has fallen behind, such as by assigning a Student Support Officer to that student. Whether or not such an intervention has occurred, and how many times an intervention has occurred, is an important variable to include. Otherwise, students who receive the intervention and are able to reach minimum standards could be classified by a model as not at risk when they in fact are at risk absent the intervention.

%%%%%%%%%%%%%%%%%%%%%%%%%%%%%%%%%%%%%%%%%%%%%%%%%%%%%%%%%%%%%%

\section{Conclusion}
\label{sec:conclusion}

\noindent
This paper applied a range of popular machine learning methods to the prediction of academic risk for Australian primary and middle school students as measured by a compulsory national standardized test. No machine learning model was able to significantly outperform standard logistic regression, despite a large data set with many predictors. 

The popularity of machine learning is not undeserved. In many areas, the increase in predictive power ML brings is very impressive. However, as we have shown in this article, it is not a panacea. Nor is the `bigness' of a data set.

%%%%%%%%%%%%%%%%%%%%%%%%%%%%%%%%%%%%%%%%%%%%%%%%%%%%%%%%%%

% Have references section title in small caps and centered
\titleformat{\section}
  {\normalfont \scshape \large \centering}
  {\thesection.}{6pt}{}

\bibliographystyle{apacite}
\bibliography{biblio}

\begin{thebibliography}{}

\bibitem [\protect \citeauthoryear {%
Bamber%
}{%
Bamber%
}{%
{\protect \APACyear {1975}}%
}]{%
BAMBER1975}
\APACinsertmetastar {%
BAMBER1975}%
\begin{APACrefauthors}%
Bamber, D.%
\end{APACrefauthors}%
\unskip\
\newblock
\APACrefYearMonthDay{1975}{}{}.
\newblock
{\BBOQ}\APACrefatitle {The area above the ordinal dominance graph and the area
  below the receiver operating characteristic graph} {The area above the
  ordinal dominance graph and the area below the receiver operating
  characteristic graph}.{\BBCQ}
\newblock
\APACjournalVolNumPages{Journal of Mathematical Psychology}{12}{4}{387 - 415}.
\newblock
\begin{APACrefDOI} \doi{https://doi.org/10.1016/0022-2496(75)90001-2}
  \end{APACrefDOI}
\PrintBackRefs{\CurrentBib}

\bibitem [\protect \citeauthoryear {%
Barro%
}{%
Barro%
}{%
{\protect \APACyear {1991}}%
}]{%
Barro1991}
\APACinsertmetastar {%
Barro1991}%
\begin{APACrefauthors}%
Barro, R\BPBI J.%
\end{APACrefauthors}%
\unskip\
\newblock
\APACrefYearMonthDay{1991}{}{}.
\newblock
{\BBOQ}\APACrefatitle {Economic Growth in a Cross Section of Countries*}
  {Economic growth in a cross section of countries*}.{\BBCQ}
\newblock
\APACjournalVolNumPages{The Quarterly Journal of Economics}{106}{2}{407-443}.
\newblock
\begin{APACrefDOI} \doi{10.2307/2937943} \end{APACrefDOI}
\PrintBackRefs{\CurrentBib}

\bibitem [\protect \citeauthoryear {%
Breiman%
}{%
Breiman%
}{%
{\protect \APACyear {1996}}%
}]{%
Breiman1996}
\APACinsertmetastar {%
Breiman1996}%
\begin{APACrefauthors}%
Breiman, L.%
\end{APACrefauthors}%
\unskip\
\newblock
\APACrefYearMonthDay{1996}{Aug}{01}.
\newblock
{\BBOQ}\APACrefatitle {Bagging predictors} {Bagging predictors}.{\BBCQ}
\newblock
\APACjournalVolNumPages{Machine Learning}{24}{2}{123--140}.
\newblock
\begin{APACrefDOI} \doi{10.1007/BF00058655} \end{APACrefDOI}
\PrintBackRefs{\CurrentBib}

\bibitem [\protect \citeauthoryear {%
Breiman%
}{%
Breiman%
}{%
{\protect \APACyear {2001}}%
}]{%
Breiman2001}
\APACinsertmetastar {%
Breiman2001}%
\begin{APACrefauthors}%
Breiman, L.%
\end{APACrefauthors}%
\unskip\
\newblock
\APACrefYearMonthDay{2001}{Oct}{01}.
\newblock
{\BBOQ}\APACrefatitle {Random Forests} {Random forests}.{\BBCQ}
\newblock
\APACjournalVolNumPages{Machine Learning}{45}{1}{5--32}.
\newblock
\begin{APACrefDOI} \doi{10.1023/A:1010933404324} \end{APACrefDOI}
\PrintBackRefs{\CurrentBib}

\bibitem [\protect \citeauthoryear {%
Breiman%
}{%
Breiman%
}{%
{\protect \APACyear {2017}}%
}]{%
Breiman2017}
\APACinsertmetastar {%
Breiman2017}%
\begin{APACrefauthors}%
Breiman, L.%
\end{APACrefauthors}%
\unskip\
\newblock
\APACrefYear{2017}.
\newblock
\APACrefbtitle {Classification and regression trees} {Classification and
  regression trees}.
\newblock
\APACaddressPublisher{}{Routledge}.
\PrintBackRefs{\CurrentBib}

\bibitem [\protect \citeauthoryear {%
Cobb-Clark%
\ \BBA {} Nguyen%
}{%
Cobb-Clark%
\ \BBA {} Nguyen%
}{%
{\protect \APACyear {2012}}%
}]{%
Cobb2012}
\APACinsertmetastar {%
Cobb2012}%
\begin{APACrefauthors}%
Cobb-Clark, D\BPBI A.%
\BCBT {}\ \BBA {} Nguyen, T\BHBI H.%
\end{APACrefauthors}%
\unskip\
\newblock
\APACrefYearMonthDay{2012}{}{}.
\newblock
{\BBOQ}\APACrefatitle {{Educational Attainment Across Generations: The Role of
  Immigration Background}} {{Educational Attainment Across Generations: The
  Role of Immigration Background}}.{\BBCQ}
\newblock
\APACjournalVolNumPages{Economic Record}{88}{283}{554--575}.
\PrintBackRefs{\CurrentBib}

\bibitem [\protect \citeauthoryear {%
Cortez%
\ \BBA {} Silva%
}{%
Cortez%
\ \BBA {} Silva%
}{%
{\protect \APACyear {2008}}%
}]{%
Cortez2008}
\APACinsertmetastar {%
Cortez2008}%
\begin{APACrefauthors}%
Cortez, P.%
\BCBT {}\ \BBA {} Silva, A\BPBI M\BPBI G.%
\end{APACrefauthors}%
\unskip\
\newblock
\APACrefYearMonthDay{2008}{}{}.
\newblock
\APACrefbtitle {Using data mining to predict secondary school student
  performance.} {Using data mining to predict secondary school student
  performance.}
\newblock
\begin{APACrefURL} \url{http://hdl.handle.net/1822/8024} \end{APACrefURL}
\PrintBackRefs{\CurrentBib}

\bibitem [\protect \citeauthoryear {%
DeLong%
, DeLong%
\BCBL {}\ \BBA {} Clarke-Pearson%
}{%
DeLong%
\ \protect \BOthers {.}}{%
{\protect \APACyear {1988}}%
}]{%
DeLong1988}
\APACinsertmetastar {%
DeLong1988}%
\begin{APACrefauthors}%
DeLong, E\BPBI R.%
, DeLong, D\BPBI M.%
\BCBL {}\ \BBA {} Clarke-Pearson, D\BPBI L.%
\end{APACrefauthors}%
\unskip\
\newblock
\APACrefYearMonthDay{1988}{}{}.
\newblock
{\BBOQ}\APACrefatitle {Comparing the Areas under Two or More Correlated
  Receiver Operating Characteristic Curves: A Nonparametric Approach}
  {Comparing the areas under two or more correlated receiver operating
  characteristic curves: A nonparametric approach}.{\BBCQ}
\newblock
\APACjournalVolNumPages{Biometrics}{44}{3}{837--845}.
\PrintBackRefs{\CurrentBib}

\bibitem [\protect \citeauthoryear {%
Ford%
}{%
Ford%
}{%
{\protect \APACyear {2013}}%
}]{%
Ford2013}
\APACinsertmetastar {%
Ford2013}%
\begin{APACrefauthors}%
Ford, M.%
\end{APACrefauthors}%
\unskip\
\newblock
\APACrefYearMonthDay{2013}{}{}.
\newblock
{\BBOQ}\APACrefatitle {{Achievement Gaps in Australia: What NAPLAN reveals
  about Education Inequality in Australia}} {{Achievement Gaps in Australia:
  What NAPLAN reveals about Education Inequality in Australia}}.{\BBCQ}
\newblock
\APACjournalVolNumPages{Race Ethnicity and Education}{16}{1}{80--102}.
\PrintBackRefs{\CurrentBib}

\bibitem [\protect \citeauthoryear {%
Friedman%
, Hastie%
\BCBL {}\ \BBA {} Tibshirani%
}{%
Friedman%
\ \protect \BOthers {.}}{%
{\protect \APACyear {2009}}%
}]{%
Friedman2001}
\APACinsertmetastar {%
Friedman2001}%
\begin{APACrefauthors}%
Friedman, J.%
, Hastie, T.%
\BCBL {}\ \BBA {} Tibshirani, R.%
\end{APACrefauthors}%
\unskip\
\newblock
\APACrefYear{2009}.
\newblock
\APACrefbtitle {The Elements of Statistical Learning} {The elements of
  statistical learning}\ (\PrintOrdinal{2}\ \BEd).
\newblock
\APACaddressPublisher{}{Springer, New York, NY}.
\newblock
\begin{APACrefDOI} \doi{https://doi.org/10.1007/978-0-387-84858-7}
  \end{APACrefDOI}
\PrintBackRefs{\CurrentBib}

\bibitem [\protect \citeauthoryear {%
Gonski%
\ \protect \BOthers {.}}{%
Gonski%
\ \protect \BOthers {.}}{%
{\protect \APACyear {2018}}%
}]{%
2gonski}
\APACinsertmetastar {%
2gonski}%
\begin{APACrefauthors}%
Gonski, D.%
, Arcus, T.%
, Boston, K.%
, Gould, V.%
, Johnson, W.%
, O'Brien, L.%
\BDBL {}Roberts, M.%
\end{APACrefauthors}%
\unskip\
\newblock
\APACrefYear{2018}.
\newblock
\APACrefbtitle {{Through Growth to Achievement: Report of the Review to Achieve
  Educational Excellence in Australian Schools}} {{Through Growth to
  Achievement: Report of the Review to Achieve Educational Excellence in
  Australian Schools}}.
\newblock
\APACaddressPublisher{Canberra}{Department of Education, Employment and
  Workplace Relations}.
\newblock
\begin{APACrefURL} \url{docs.education.gov.au/} \end{APACrefURL}
\PrintBackRefs{\CurrentBib}

\bibitem [\protect \citeauthoryear {%
Gray%
, McGuinness%
\BCBL {}\ \BBA {} Owende%
}{%
Gray%
\ \protect \BOthers {.}}{%
{\protect \APACyear {2013}}%
}]{%
Gray2013}
\APACinsertmetastar {%
Gray2013}%
\begin{APACrefauthors}%
Gray, G.%
, McGuinness, C.%
\BCBL {}\ \BBA {} Owende, P.%
\end{APACrefauthors}%
\unskip\
\newblock
\APACrefYearMonthDay{2013}{}{}.
\newblock
{\BBOQ}\APACrefatitle {An investigation of psychometric measures for modelling
  academic performance in tertiary education} {An investigation of psychometric
  measures for modelling academic performance in tertiary education}.{\BBCQ}
\newblock
\BIn{} \APACrefbtitle {Educational Data Mining 2013.} {Educational data mining
  2013.}
\PrintBackRefs{\CurrentBib}

\bibitem [\protect \citeauthoryear {%
Hoerl%
\ \BBA {} Kennard%
}{%
Hoerl%
\ \BBA {} Kennard%
}{%
{\protect \APACyear {1970}}%
}]{%
Hoerl1970}
\APACinsertmetastar {%
Hoerl1970}%
\begin{APACrefauthors}%
Hoerl, A\BPBI E.%
\BCBT {}\ \BBA {} Kennard, R\BPBI W.%
\end{APACrefauthors}%
\unskip\
\newblock
\APACrefYearMonthDay{1970}{}{}.
\newblock
{\BBOQ}\APACrefatitle {Ridge Regression: Biased Estimation for Nonorthogonal
  Problems} {Ridge regression: Biased estimation for nonorthogonal
  problems}.{\BBCQ}
\newblock
\APACjournalVolNumPages{Technometrics}{12}{1}{55-67}.
\newblock
\begin{APACrefDOI} \doi{10.1080/00401706.1970.10488634} \end{APACrefDOI}
\PrintBackRefs{\CurrentBib}

\bibitem [\protect \citeauthoryear {%
Huang%
\ \BBA {} Fang%
}{%
Huang%
\ \BBA {} Fang%
}{%
{\protect \APACyear {2013}}%
}]{%
HUANG2013}
\APACinsertmetastar {%
HUANG2013}%
\begin{APACrefauthors}%
Huang, S.%
\BCBT {}\ \BBA {} Fang, N.%
\end{APACrefauthors}%
\unskip\
\newblock
\APACrefYearMonthDay{2013}{}{}.
\newblock
{\BBOQ}\APACrefatitle {Predicting student academic performance in an
  engineering dynamics course: A comparison of four types of predictive
  mathematical models} {Predicting student academic performance in an
  engineering dynamics course: A comparison of four types of predictive
  mathematical models}.{\BBCQ}
\newblock
\APACjournalVolNumPages{Computers \& Education}{61}{}{133 - 145}.
\newblock
\begin{APACrefDOI} \doi{10.1016/j.compedu.2012.08.015} \end{APACrefDOI}
\PrintBackRefs{\CurrentBib}

\bibitem [\protect \citeauthoryear {%
Jishan%
, Rashu%
, Haque%
\BCBL {}\ \BBA {} Rahman%
}{%
Jishan%
\ \protect \BOthers {.}}{%
{\protect \APACyear {2015}}%
}]{%
Jishan2015}
\APACinsertmetastar {%
Jishan2015}%
\begin{APACrefauthors}%
Jishan, S\BPBI T.%
, Rashu, R\BPBI I.%
, Haque, N.%
\BCBL {}\ \BBA {} Rahman, R\BPBI M.%
\end{APACrefauthors}%
\unskip\
\newblock
\APACrefYearMonthDay{2015}{Mar}{12}.
\newblock
{\BBOQ}\APACrefatitle {Improving accuracy of students' final grade prediction
  model using optimal equal width binning and synthetic minority over-sampling
  technique} {Improving accuracy of students' final grade prediction model
  using optimal equal width binning and synthetic minority over-sampling
  technique}.{\BBCQ}
\newblock
\APACjournalVolNumPages{Decision Analytics}{2}{1}{1}.
\newblock
\begin{APACrefDOI} \doi{10.1186/s40165-014-0010-2} \end{APACrefDOI}
\PrintBackRefs{\CurrentBib}

\bibitem [\protect \citeauthoryear {%
S.~Kotsiantis%
, Pierrakeas%
\BCBL {}\ \BBA {} Pintelas%
}{%
S.~Kotsiantis%
\ \protect \BOthers {.}}{%
{\protect \APACyear {2004}}%
}]{%
Kotsiantis2004}
\APACinsertmetastar {%
Kotsiantis2004}%
\begin{APACrefauthors}%
Kotsiantis, S.%
, Pierrakeas, C.%
\BCBL {}\ \BBA {} Pintelas, P.%
\end{APACrefauthors}%
\unskip\
\newblock
\APACrefYearMonthDay{2004}{}{}.
\newblock
{\BBOQ}\APACrefatitle {PREDICTING STUDENTS' PERFORMANCE IN DISTANCE LEARNING
  USING MACHINE LEARNING TECHNIQUES} {Predicting students' performance in
  distance learning using machine learning techniques}.{\BBCQ}
\newblock
\APACjournalVolNumPages{Applied Artificial Intelligence}{18}{5}{411-426}.
\newblock
\begin{APACrefDOI} \doi{10.1080/08839510490442058} \end{APACrefDOI}
\PrintBackRefs{\CurrentBib}

\bibitem [\protect \citeauthoryear {%
S\BPBI B.~Kotsiantis%
}{%
S\BPBI B.~Kotsiantis%
}{%
{\protect \APACyear {2012}}%
}]{%
Kotsiantis2012}
\APACinsertmetastar {%
Kotsiantis2012}%
\begin{APACrefauthors}%
Kotsiantis, S\BPBI B.%
\end{APACrefauthors}%
\unskip\
\newblock
\APACrefYearMonthDay{2012}{Apr}{01}.
\newblock
{\BBOQ}\APACrefatitle {Use of machine learning techniques for educational
  proposes: a decision support system for forecasting students' grades} {Use of
  machine learning techniques for educational proposes: a decision support
  system for forecasting students' grades}.{\BBCQ}
\newblock
\APACjournalVolNumPages{Artificial Intelligence Review}{37}{4}{331--344}.
\newblock
\begin{APACrefDOI} \doi{10.1007/s10462-011-9234-x} \end{APACrefDOI}
\PrintBackRefs{\CurrentBib}

\bibitem [\protect \citeauthoryear {%
Li%
, Jia%
\BCBL {}\ \BBA {} Zhao%
}{%
Li%
\ \protect \BOthers {.}}{%
{\protect \APACyear {2013}}%
}]{%
Li2013}
\APACinsertmetastar {%
Li2013}%
\begin{APACrefauthors}%
Li, J.%
, Jia, Y.%
\BCBL {}\ \BBA {} Zhao, Z.%
\end{APACrefauthors}%
\unskip\
\newblock
\APACrefYearMonthDay{2013}{May}{01}.
\newblock
{\BBOQ}\APACrefatitle {Partly adaptive elastic net and its application to
  microarray classification} {Partly adaptive elastic net and its application
  to microarray classification}.{\BBCQ}
\newblock
\APACjournalVolNumPages{Neural Computing and Applications}{22}{6}{1193--1200}.
\newblock
\begin{APACrefDOI} \doi{10.1007/s00521-012-0885-6} \end{APACrefDOI}
\PrintBackRefs{\CurrentBib}

\bibitem [\protect \citeauthoryear {%
Li%
\ \BBA {} Jia%
}{%
Li%
\ \BBA {} Jia%
}{%
{\protect \APACyear {2010}}%
}]{%
Li2010}
\APACinsertmetastar {%
Li2010}%
\begin{APACrefauthors}%
Li, J.%
\BCBT {}\ \BBA {} Jia, Y\BHBI M.%
\end{APACrefauthors}%
\unskip\
\newblock
\APACrefYearMonthDay{2010}{}{}.
\newblock
{\BBOQ}\APACrefatitle {An Improved Elastic Net for Cancer Classification and
  Gene Selection} {An improved elastic net for cancer classification and gene
  selection}.{\BBCQ}
\newblock
\APACjournalVolNumPages{Acta Automatica Sinica}{36}{7}{976 - 981}.
\newblock
\begin{APACrefDOI} \doi{https://doi.org/10.1016/S1874-1029(09)60042-2}
  \end{APACrefDOI}
\PrintBackRefs{\CurrentBib}

\bibitem [\protect \citeauthoryear {%
Nicoletti%
\ \BBA {} Rabe%
}{%
Nicoletti%
\ \BBA {} Rabe%
}{%
{\protect \APACyear {2013}}%
}]{%
Nicoletti2013}
\APACinsertmetastar {%
Nicoletti2013}%
\begin{APACrefauthors}%
Nicoletti, C.%
\BCBT {}\ \BBA {} Rabe, B.%
\end{APACrefauthors}%
\unskip\
\newblock
\APACrefYearMonthDay{2013}{}{}.
\newblock
{\BBOQ}\APACrefatitle {Inequality in Pupils' Test Scores: How Much Do Family,
  Sibling Type and Neighbourhood Matter?} {Inequality in pupils' test scores:
  How much do family, sibling type and neighbourhood matter?}{\BBCQ}
\newblock
\APACjournalVolNumPages{Economica}{80}{318}{197 - 218}.
\PrintBackRefs{\CurrentBib}

\bibitem [\protect \citeauthoryear {%
Noether%
}{%
Noether%
}{%
{\protect \APACyear {1967}}%
}]{%
noether1967}
\APACinsertmetastar {%
noether1967}%
\begin{APACrefauthors}%
Noether, G\BPBI E.%
\end{APACrefauthors}%
\unskip\
\newblock
\APACrefYearMonthDay{1967}{}{}.
\newblock
\APACrefbtitle {Elements of nonparametric statistics} {Elements of
  nonparametric statistics}\ \APACbVolEdTR{}{\BTR{}}.
\newblock
\APACaddressInstitution{}{Wiley \& Sons}.
\PrintBackRefs{\CurrentBib}

\bibitem [\protect \citeauthoryear {%
Reback%
}{%
Reback%
}{%
{\protect \APACyear {2008}}%
}]{%
Reback2008}
\APACinsertmetastar {%
Reback2008}%
\begin{APACrefauthors}%
Reback, R.%
\end{APACrefauthors}%
\unskip\
\newblock
\APACrefYearMonthDay{2008}{}{}.
\newblock
{\BBOQ}\APACrefatitle {{Teaching to the rating: School accountability and the
  distribution of student achievement}} {{Teaching to the rating: School
  accountability and the distribution of student achievement}}.{\BBCQ}
\newblock
\APACjournalVolNumPages{Journal of Public Economics}{92}{5}{1394 - 1415}.
\PrintBackRefs{\CurrentBib}

\bibitem [\protect \citeauthoryear {%
Romero%
\ \BBA {} Ventura%
}{%
Romero%
\ \BBA {} Ventura%
}{%
{\protect \APACyear {2007}}%
}]{%
Romero2007}
\APACinsertmetastar {%
Romero2007}%
\begin{APACrefauthors}%
Romero, C.%
\BCBT {}\ \BBA {} Ventura, S.%
\end{APACrefauthors}%
\unskip\
\newblock
\APACrefYearMonthDay{2007}{}{}.
\newblock
{\BBOQ}\APACrefatitle {Educational data mining: A survey from 1995 to 2005}
  {Educational data mining: A survey from 1995 to 2005}.{\BBCQ}
\newblock
\APACjournalVolNumPages{Expert Systems with Applications}{33}{1}{135 - 146}.
\newblock
\begin{APACrefDOI} \doi{https://doi.org/10.1016/j.eswa.2006.04.005}
  \end{APACrefDOI}
\PrintBackRefs{\CurrentBib}

\bibitem [\protect \citeauthoryear {%
Rumelhart%
, Hinton%
\BCBL {}\ \BBA {} Williams%
}{%
Rumelhart%
\ \protect \BOthers {.}}{%
{\protect \APACyear {1985}}%
}]{%
Rumelhart1985}
\APACinsertmetastar {%
Rumelhart1985}%
\begin{APACrefauthors}%
Rumelhart, D\BPBI E.%
, Hinton, G\BPBI E.%
\BCBL {}\ \BBA {} Williams, R\BPBI J.%
\end{APACrefauthors}%
\unskip\
\newblock
\APACrefYearMonthDay{1985}{}{}.
\newblock
\APACrefbtitle {Learning internal representations by error propagation}
  {Learning internal representations by error propagation}\
  \APACbVolEdTR{}{\BTR{}}.
\newblock
\APACaddressInstitution{}{California Univ San Diego La Jolla Inst for Cognitive
  Science}.
\PrintBackRefs{\CurrentBib}

\bibitem [\protect \citeauthoryear {%
Shingari%
, Kumar%
\BCBL {}\ \BBA {} Khetan%
}{%
Shingari%
\ \protect \BOthers {.}}{%
{\protect \APACyear {2017}}%
}]{%
Shingari2017}
\APACinsertmetastar {%
Shingari2017}%
\begin{APACrefauthors}%
Shingari, I.%
, Kumar, D.%
\BCBL {}\ \BBA {} Khetan, M.%
\end{APACrefauthors}%
\unskip\
\newblock
\APACrefYearMonthDay{2017}{}{}.
\newblock
{\BBOQ}\APACrefatitle {A review of applications of data mining techniques for
  prediction of students’ performance in higher education} {A review of
  applications of data mining techniques for prediction of students’
  performance in higher education}.{\BBCQ}
\newblock
\APACjournalVolNumPages{Journal of Statistics and Management
  Systems}{20}{4}{713-722}.
\newblock
\begin{APACrefDOI} \doi{10.1080/09720510.2017.1395191} \end{APACrefDOI}
\PrintBackRefs{\CurrentBib}

\bibitem [\protect \citeauthoryear {%
Srivastava%
, Hinton%
, Krizhevsky%
, Sutskever%
\BCBL {}\ \BBA {} Salakhutdinov%
}{%
Srivastava%
\ \protect \BOthers {.}}{%
{\protect \APACyear {2014}}%
}]{%
Srivastava2014}
\APACinsertmetastar {%
Srivastava2014}%
\begin{APACrefauthors}%
Srivastava, N.%
, Hinton, G.%
, Krizhevsky, A.%
, Sutskever, I.%
\BCBL {}\ \BBA {} Salakhutdinov, R.%
\end{APACrefauthors}%
\unskip\
\newblock
\APACrefYearMonthDay{2014}{}{}.
\newblock
{\BBOQ}\APACrefatitle {Dropout: A Simple Way to Prevent Neural Networks from
  Overfitting} {Dropout: A simple way to prevent neural networks from
  overfitting}.{\BBCQ}
\newblock
\APACjournalVolNumPages{Journal of Machine Learning Research}{15}{}{1929-1958}.
\newblock
\begin{APACrefURL} \url{http://jmlr.org/papers/v15/srivastava14a.html}
  \end{APACrefURL}
\PrintBackRefs{\CurrentBib}

\bibitem [\protect \citeauthoryear {%
Tibshirani%
}{%
Tibshirani%
}{%
{\protect \APACyear {1996}}%
}]{%
Tibshirani1996}
\APACinsertmetastar {%
Tibshirani1996}%
\begin{APACrefauthors}%
Tibshirani, R.%
\end{APACrefauthors}%
\unskip\
\newblock
\APACrefYearMonthDay{1996}{}{}.
\newblock
{\BBOQ}\APACrefatitle {Regression Shrinkage and Selection Via the Lasso}
  {Regression shrinkage and selection via the lasso}.{\BBCQ}
\newblock
\APACjournalVolNumPages{Journal of the Royal Statistical Society: Series B
  (Methodological)}{58}{1}{267-288}.
\newblock
\begin{APACrefDOI} \doi{10.1111/j.2517-6161.1996.tb02080.x} \end{APACrefDOI}
\PrintBackRefs{\CurrentBib}

\bibitem [\protect \citeauthoryear {%
Vandamme%
, Meskens%
\BCBL {}\ \BBA {} Superby%
}{%
Vandamme%
\ \protect \BOthers {.}}{%
{\protect \APACyear {2007}}%
}]{%
Vandamme2007}
\APACinsertmetastar {%
Vandamme2007}%
\begin{APACrefauthors}%
Vandamme, J.%
, Meskens, N.%
\BCBL {}\ \BBA {} Superby, J.%
\end{APACrefauthors}%
\unskip\
\newblock
\APACrefYearMonthDay{2007}{}{}.
\newblock
{\BBOQ}\APACrefatitle {Predicting Academic Performance by Data Mining Methods}
  {Predicting academic performance by data mining methods}.{\BBCQ}
\newblock
\APACjournalVolNumPages{Education Economics}{15}{4}{405-419}.
\newblock
\begin{APACrefDOI} \doi{10.1080/09645290701409939} \end{APACrefDOI}
\PrintBackRefs{\CurrentBib}

\bibitem [\protect \citeauthoryear {%
Wan%
, Zeiler%
, Zhang%
, Cun%
\BCBL {}\ \BBA {} Fergus%
}{%
Wan%
\ \protect \BOthers {.}}{%
{\protect \APACyear {2013}}%
}]{%
Wan2013}
\APACinsertmetastar {%
Wan2013}%
\begin{APACrefauthors}%
Wan, L.%
, Zeiler, M.%
, Zhang, S.%
, Cun, Y\BPBI L.%
\BCBL {}\ \BBA {} Fergus, R.%
\end{APACrefauthors}%
\unskip\
\newblock
\APACrefYearMonthDay{2013}{17--19 Jun}{}.
\newblock
{\BBOQ}\APACrefatitle {Regularization of Neural Networks using DropConnect}
  {Regularization of neural networks using dropconnect}.{\BBCQ}
\newblock
\BIn{} S.~Dasgupta\ \BBA {} D.~McAllester\ (\BEDS), \APACrefbtitle {Proceedings
  of the 30th International Conference on Machine Learning} {Proceedings of the
  30th international conference on machine learning}\ (\BVOL~28, \BPGS\
  1058--1066).
\newblock
\APACaddressPublisher{Atlanta, Georgia, USA}{PMLR}.
\newblock
\begin{APACrefURL} \url{http://proceedings.mlr.press/v28/wan13.html}
  \end{APACrefURL}
\PrintBackRefs{\CurrentBib}

\bibitem [\protect \citeauthoryear {%
Yadav%
\ \BBA {} Pal%
}{%
Yadav%
\ \BBA {} Pal%
}{%
{\protect \APACyear {2012}}%
}]{%
Yadav2012}
\APACinsertmetastar {%
Yadav2012}%
\begin{APACrefauthors}%
Yadav, S\BPBI K.%
\BCBT {}\ \BBA {} Pal, S.%
\end{APACrefauthors}%
\unskip\
\newblock
\APACrefYearMonthDay{2012}{}{}.
\newblock
{\BBOQ}\APACrefatitle {Data mining: A prediction for performance improvement of
  engineering students using classification} {Data mining: A prediction for
  performance improvement of engineering students using classification}.{\BBCQ}
\newblock
\APACjournalVolNumPages{arXiv preprint arXiv:1203.3832}{}{}{}.
\PrintBackRefs{\CurrentBib}

\bibitem [\protect \citeauthoryear {%
Zou%
\ \BBA {} Hastie%
}{%
Zou%
\ \BBA {} Hastie%
}{%
{\protect \APACyear {2005}}%
}]{%
Zou2005}
\APACinsertmetastar {%
Zou2005}%
\begin{APACrefauthors}%
Zou, H.%
\BCBT {}\ \BBA {} Hastie, T.%
\end{APACrefauthors}%
\unskip\
\newblock
\APACrefYearMonthDay{2005}{}{}.
\newblock
{\BBOQ}\APACrefatitle {Regularization and variable selection via the elastic
  net} {Regularization and variable selection via the elastic net}.{\BBCQ}
\newblock
\APACjournalVolNumPages{Journal of the Royal Statistical Society: Series B
  (Statistical Methodology)}{67}{2}{301-320}.
\newblock
\begin{APACrefDOI} \doi{10.1111/j.1467-9868.2005.00503.x} \end{APACrefDOI}
\PrintBackRefs{\CurrentBib}

\end{thebibliography}
\onecolumn
%%%%%%%%%%%%%%%%%%%%%%%%%%%%%%%%%%%%%%%%%%%%%%%%%%%%%%%%%

% Tables and figures

\clearpage
%%%%%%%%%%%%%%%%%%%%%%%%%%%%%%%%%%%%

%%%%%%%%%%%%%%%%%%%%%%%%%%%%%

% Decision trees. Weighted.
%\newgeometry{left=0.5in, right=0.5in}

\clearpage

%%%%%%%%%%%%%%%%%%%%%%%%%%%%%%%%%%%%%%%%%%%%%%%%%% %%%%%%%%%%%%%%%%%%%%%%%%%%%%%

%%%%%%%%%%%%%%%
% Variable Description Table

% Have to mess with the line spread to make the table 
% and footnotes fit on same page.

\renewcommand{\arraystretch}{1.3}
\begin{table}[!ph]
\centering
\begin{threeparttable}
\caption{Variables contained in the NAPLAN data set}
\label{tab:variablelist}
\footnotesize
    \begin{tabular}{ l  | p{9.2cm} } \toprule
    \textbf{Variable} & \textbf{Description} \\ \midrule
    Reading standard [\textit{readrisk}] & =1 if student is below standard in reading, and =0 if at standard \\

    Math standard [\textit{numrisk}] & =1 if student is below standard in numeracy, and =0 if at standard \\ 

    Previous Reading standard [\textit{prevreadrisk}] & =1 if student was below standard in reading on previous NAPLAN, and =0 if at standard \\

    Previous Math standard [\textit{prevnumrisk}] &  =1 if student was below standard in numeracy on previous NAPLAN, and =0 if at standard \\

    Private schooling [\textit{private}] & =1 if student attends a non-government school, and =0 if government school \\

    Age [\textit{age}] & Age at the time of taking the test to one decimal place  \\

    LBOTE [\textit{LBOTE}]& =1 if the student has a language background other than English, and =0 if not \\

    Indigenous [\textit{indigenous}]& =1 if the student identifies as Indigenous Australian or Torres Strait Islander, and =0 if not \\

    Female [\textit{female}]& =1 if the student is female, and =0 if male \\

    State [\textit{state}] & Categorical variable with levels: \textit{South Australia}, \textit{New South Wales}, \textit{Victoria}, \textit{Queensland}, \textit{Tasmania}, \textit{Western Australia},  \textit{Australian Capital Territory}, and \textit{Northern Territory};  denoting the state in which the student resides.\\

    Mother's education [\textit{mumschool}] &Categorical variable =1 if mother completed up to grade 9,  =2 for grade 10, =3 for grade 11, and =4 for grade 12.\\

    Mother's higher education [\textit{mumhighed}]& Categorical variable, =5 if mother's highest level of education is a certificate I to IV or trade qualification, =6 if a diploma or advanced diploma, =7 if bachelor's degree, =8 if none.\\

    Mother's employment [\textit{mumoccup}] & Categorical variable, =1 if mother employed in category 1,
\tablefootnote{Category 1: Senior management in a large business organisation, government administration, or defence, and qualified professionals; \textit{e.g.} business/policy analyst, defence forces commissioned officer, professionals with degree or higher qualifications, administrators such as school principals, \textit{etc.}}
=2 if category 2,
\tablefootnote{Category 2: Other business managers, arts/media/sportspersons, and associate professionals; \textit{e.g.} owner/manager of a farm or business, retail sales/service manager, musician, journalist, designer, sports official, business/administrative staff, \textit{etc.}}
=3 if category 3,
\tablefootnote{Category 3: Tradespeople, clerks, and skilled office, sales, and service staff; \textit{e.g.} 4 year trade certificate by apprenticeship, clerks, personal assistants, sales, flight attendants, fitness instructors, child care workers, \textit{etc.}}
=4 if category 4,
\tablefootnote{Category 4: Machine operators, hospitality staff, assistants, labourers and related workers; \textit{e.g.} machine operators, drivers, labourers, office assistants, defence forces ranked below senior non-commissioned officer, miners, farmers, factory hands, guards, \textit{etc.}}
=8 if unemployed\\

      Father's education [\textit{dadschool}] & Categorical variable,  =1 if father completed up to grade 9, =2 if grade 10, =3 if grade 11, and =4 if grade 12\\

    Father's higher education [\textit{dadhighed}] &Categorical variable, =5 if father's highest level of education is a certificate I to IV or trade qualification, =6 if a diploma or advanced diploma, =7 if bachelor's degree, =8 if none.\\

        Father's employment [\textit{dadoccup}] &Categorical variable, =1 if father employed in category 1, =2 if category 2, = 3 if category 3, =4 if category 4, =8 if unemployed\\
    \bottomrule
    \end{tabular}
\fignote{\scriptsize Raw data provided by the Australian Curriculum, Assessment and Reporting Authority through their Data Access Program. Variable name codes given in square brackets.}
\end{threeparttable}
\end{table}
\clearpage
\linespread{1}
%%%%%%%%%%%%%%%%%%%%%%%%%%%%%%%%%%

%%%%%%%%%%%%%%%%%%%%%%%%%%

\end{document}